\documentclass[default,iicol]{sn-jnl}

\jyear{2021}%

\theoremstyle{thmstyleone}%
\theoremstyle{thmstyletwo}%

\theoremstyle{thmstylethree}%

\usepackage{subcaption}

\usepackage{colortbl}
\usepackage{color, xcolor} 
\usepackage[table]{xcolor}

\DeclareRobustCommand{\gain}[1]{~\scalebox{.70}{\textcolor{teal}{(+#1)}}}

\usepackage{array}
\usepackage{ragged2e}
\usepackage{threeparttable}

\usepackage{tabularx}
\usepackage{makecell}
\usepackage{booktabs}

\newcolumntype{L}[1]{>{\RaggedRight\arraybackslash}p{#1}}
\definecolor{row}{RGB}{235, 245, 251}

\def\eg{\emph{e.g.}}

\begin{document}

\title[Video Understanding Survey]{Video Understanding: From Geometry and Semantics to Unified Models}


\author[1]{\fnm{Zhaochong} \sur{An}}

\author[2]{\fnm{Zirui} \sur{Li}}

\author[3]{\fnm{Mingqiao} \sur{Ye}}

\author[4]{\fnm{Feng} \sur{Qiao}}

\author[1]{\fnm{Jiaang} \sur{Li}}

\author[5]{\fnm{Zongwei} \sur{Wu}}

\author[6,7]{\fnm{Vishal} \sur{Thengane}}

\author[1,8]{\fnm{Chengzu} \sur{Li}}

\author[9]{\fnm{Lei} \sur{Li}}

\author[10]{\fnm{Luc} \sur{Van Gool}}

\author[2]{\fnm{Guolei} \sur{Sun}}
\equalcont{Corresponding author.}

\author[1]{\fnm{Serge} \sur{Belongie}}

\affil[1]{\orgdiv{Department of Computer Science}, \orgname{University of Copenhagen}, \orgaddress{\city{Copenhagen} \postcode{2100},  \country{Denmark}}}

\affil[2]{\orgdiv{College of Computer Science}, \orgname{Nankai University}, \orgaddress{\city{Tianjin} \postcode{300350},  \country{China}}}

\affil[3]{\orgdiv{School of Computer and Communication Sciences}, \orgname{EPFL}, \orgaddress{\city{Lausanne} \postcode{1015},  \country{Switzerland}}}

\affil[4]{\orgdiv{Department of Computer Science \& Engineering}, \orgname{Washington University in St. Louis}, \orgaddress{\city{St. Louis} \postcode{MO 63130},  \country{USA}}}

\affil[5]{\orgdiv{Computer Vision Lab}, \orgname{University of Würzburg}, \orgaddress{\city{Würzburg} \postcode{97070},  \country{Germany}}}

\affil[6]{\orgdiv{Computer Science Research Centre}, \orgname{University of Surrey}, \orgaddress{\city{Guildford} \postcode{GU2 7XH},  \country{UK}}}

\affil[7]{\orgdiv{School of Electrical, Computer and Telecommunications Engineering }, 
\orgname{University of Wollongong}, \orgaddress{\city{Wollongong} \postcode{2500},  \country{Australia}}}

\affil[8]{\orgdiv{Language Technology Lab}, \orgname{University of Cambridge}, \orgaddress{\city{Cambridge} \postcode{CB2 1TN},  \country{UK}}}

\affil[9]{\orgdiv{School of Artificial Intelligence}, \orgname{Beijing Institute of Technology}, \orgaddress{\city{Beijing} \postcode{100081},  \country{China}}}

\affil[10]{\orgdiv{Institute for Computer Science}, \orgname{INSAIT}, \orgaddress{\city{Sofia} \postcode{1784},  \country{Bulgaria}}}



\abstract{
Video understanding aims to enable models to perceive, reason about, and interact with the dynamic visual world. 
In contrast to image understanding, video understanding inherently requires modeling temporal dynamics and evolving visual context, placing stronger demands on spatiotemporal reasoning and making it a foundational problem in computer vision.
In this survey, we present a structured overview of video understanding by organizing the literature into three complementary perspectives: \emph{low-level video geometry understanding}, \emph{high-level semantic understanding}, and \emph{unified video understanding models}. 
We further highlight a broader shift from \emph{isolated, task-specific} pipelines toward \emph{unified modeling} paradigms that can be adapted to diverse downstream objectives, enabling a more systematic view of recent progress. 
By consolidating these perspectives, this survey provides a coherent map of the evolving video understanding landscape, summarizes key modeling trends and design principles, and outlines open challenges toward building robust, scalable, and unified video foundation models.
}




\maketitle

\begin{figure*}[t]
    \centering
    \setlength{\abovecaptionskip}{2mm}
    \includegraphics[width=0.999\linewidth]{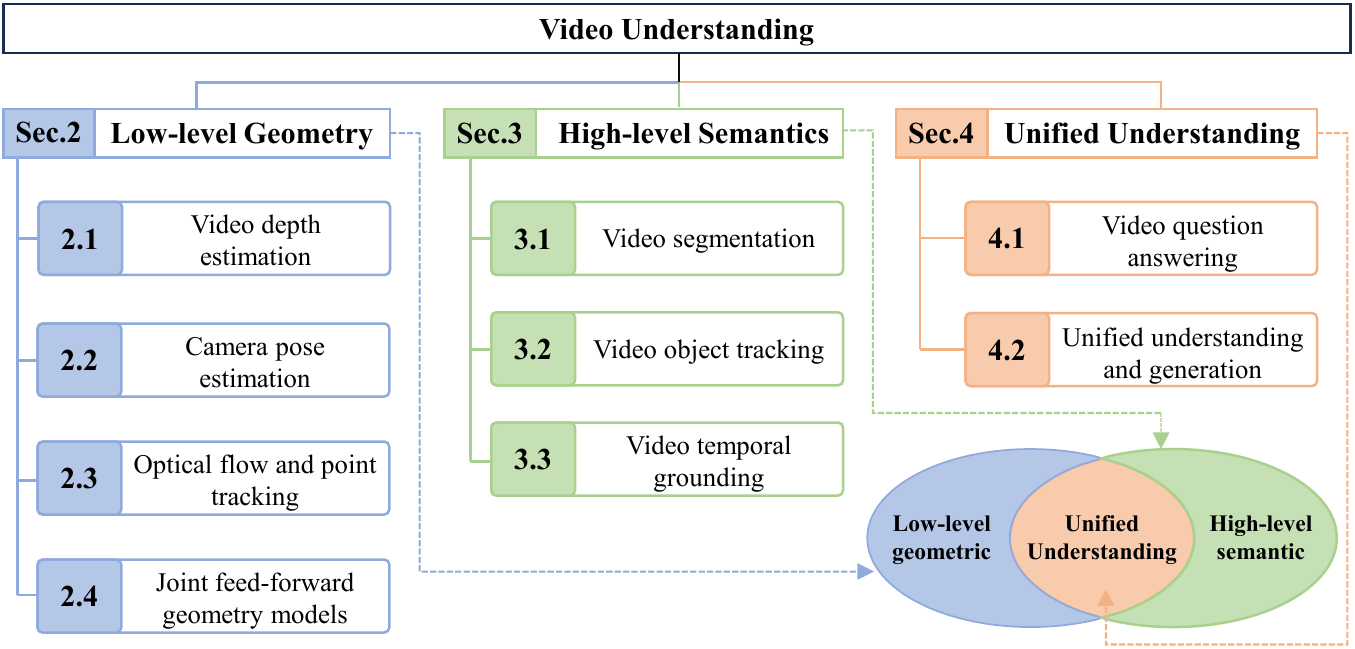}
\caption{\textbf{Classification of video understanding by level.} 
This survey organizes video understanding methods into three categories: 
(1) low-level geometry understanding, 
(2) high-level semantic understanding, and 
(3) unified video understanding. 
The right diagram illustrates the conceptual relationship among the three levels.
}
\label{fig:framework}
\end{figure*}

\section{Introduction}
\label{sec:introduction}

Video understanding is a fundamental task for enabling models to perceive and reason about the surrounding world. 
In contrast to static images, videos exhibit temporal continuity, viewpoint changes, motion-induced occlusions, and long-range dependencies across frames~\cite{li2026thinking,yang2025vca,ding2025videozoomer,zhao2025video,an2025onestory,gao2025ditvr}, which substantially increase the difficulty of the task. 
From a conceptual perspective, video understanding can be viewed at two complementary levels. 
At the \emph{low level}, models aim to recover \textbf{geometric structure}, such as depth, camera motion, and point correspondences, in order to describe \emph{how the world moves}~\cite{wang2025vggt,lin2025depth,liu2025scaling}. 
At the \emph{high level}, models seek to capture \textbf{semantic information}, including objects, actions, interactions, and language-grounded concepts, which convey \emph{what the world means}~\cite{li2025sam3,tan2025you,qin2025question}. 
Recent advances in large-scale pretraining and transformer-based architectures~\cite{wang2024dust3r,ho2022video,wu2024next} have led to significant improvements at both levels, enabling increasingly robust video understanding systems.

A key observation motivating this survey is that progress in video understanding has largely evolved along three intertwined directions: recovering geometric structure, recognizing and reasoning about semantics over time, and building unified models that integrate both.
Many classical pipelines and modern learned systems~\cite{kopf2021rcvd,wang2023tracking,ravi2024sam2} emphasize either geometry or semantics, while an emerging class of unified models requires both simultaneously.
In such settings, geometric modeling supports underlying physical and temporal consistency, while semantic understanding preserves object identity and compositional meaning across long video sequences~\cite{wiedemer2025video,tong2025thinking,liu2025can,guo2025video}. 
Accordingly, this survey is organized around these three complementary perspectives, which together provide a coherent view of the evolving landscape of video understanding, as summarized in Fig.~\ref{fig:framework}.

\paragraph{Low-level video geometry understanding.}
The first part of the survey focuses on \textbf{low-level geometric understanding from video}, where the goal is to predict physically grounded representations from video inputs~\cite{wang2024dust3r,depth_anything_v1,zholus2025tapnext}. Representative outputs include depth, camera pose, dense optical flow, and long-term point tracking across frames. 
These geometric predictions serve as a foundation for scene reconstruction and geometry-aware reasoning, providing a natural bridge between 2D visual observations and underlying 3D scene structure~\cite{hu2025vggt4d,liu2025scaling,wiedemer2025video}. 
In addition to task-specific predictors, we highlight a rapidly growing family of \emph{joint feed-forward geometry models}~\cite{wang2025vggt,lin2025depth} that learn geometry in a data-driven manner and jointly predict multiple geometric primitives within a unified network, enabling improved generalization across scenes and video domains.

\paragraph{High-level semantic understanding.}
The second part of the survey reviews \textbf{high-level semantic understanding of video}, where the goal is to recognize, localize, and reason about meaningful entities and events over time~\cite{li2025sam3,tan2025you,wang2025time,zhou2025steering}. 
We focus on three representative task families: video segmentation, video object tracking, and video temporal grounding. 
Although these tasks differ in formulation and supervision, they share a common requirement for temporally coherent representations that preserve object identity, capture motion cues, and integrate contextual information across frames. 
We discuss how modern architectures and large-scale pretrained models improve robustness, for example, by incorporating scalable designs~\cite{li2025sam3,ravi2024sam2,an2023temporal,cheng2022masked}, additional modalities~\cite{untrack,tan2025you,zhou2024event,zhou2023rgb}, and large language or vision--language models~\cite{chen2023grounding,qin2025question}.

\paragraph{Unified video understanding models.}
The third part of the survey focuses on models that explicitly aim for \textbf{unified video understanding}, requiring the simultaneous handling of low-level geometry and high-level semantics. 
This setting arises naturally when a system must \emph{answer questions about} a video rather than merely assign labels. 
For example, \textbf{video question answering} models must reason over both geometric cues (\eg, depth, scene structure, and correspondences) and semantic information (\eg, object identity, intent, and interactions) to produce correct and explainable responses. 
Similarly, emerging \textbf{unified understanding--generation} models~\cite{xie2025showo2,deng2025emerging,liu2025tuna} combine video synthesis capabilities, which require fine-grained spatiotemporal and geometric consistency, with understanding-oriented interfaces such as instruction following and video editing that depend on semantic interpretation. 
We argue that these unified models provide a practical lens for next-generation video foundation models, as real-world applications increasingly demand both geometry-consistent generation~\cite{kupyn2025epipolar,zhang2025dualcamctrl,wang2025physcorr} and compositional semantic control~\cite{zhou2025scaling,liu2025stablev2v,gao2025lora} within a single system.

\paragraph{Survey structure.}
Following this organization, the survey provides a coherent map of the field, progressing from geometric primitives to semantic reasoning and finally to models that integrate both. 
We first cover \textbf{low-level video geometry understanding} (Section~\ref{sec:low_level_geometry}), including video depth estimation (Section~\ref{sec:video_depth}), camera pose estimation (Section~\ref{sec:camera_pose_estimation}), optical flow and point tracking (Section~\ref{sec:point_tracking}), and joint feed-forward geometry models (Section~\ref{sec:feedforward_geometry}). 
We then review \textbf{high-level semantic understanding} (Section~\ref{sec:high_level_semantics}), focusing on video segmentation (Section~\ref{sec:video_segmentation}), video object tracking (Section~\ref{sec:video_object_tracking}), and video temporal grounding (Section~\ref{sec:video_grounding}). 
Finally, we discuss \textbf{unified video understanding models} (Section~\ref{sec:unified}), covering video question answering (Section~\ref{sec:vqa}) and unified understanding--generation systems (Section~\ref{sec:vug}). 
We conclude with open challenges and future directions (Section~\ref{sec:conclusion}).

\section{Low-level Video Geometry Understanding}
\label{sec:low_level_geometry}

\begin{figure*}
    \centering
    \includegraphics[width=\linewidth]{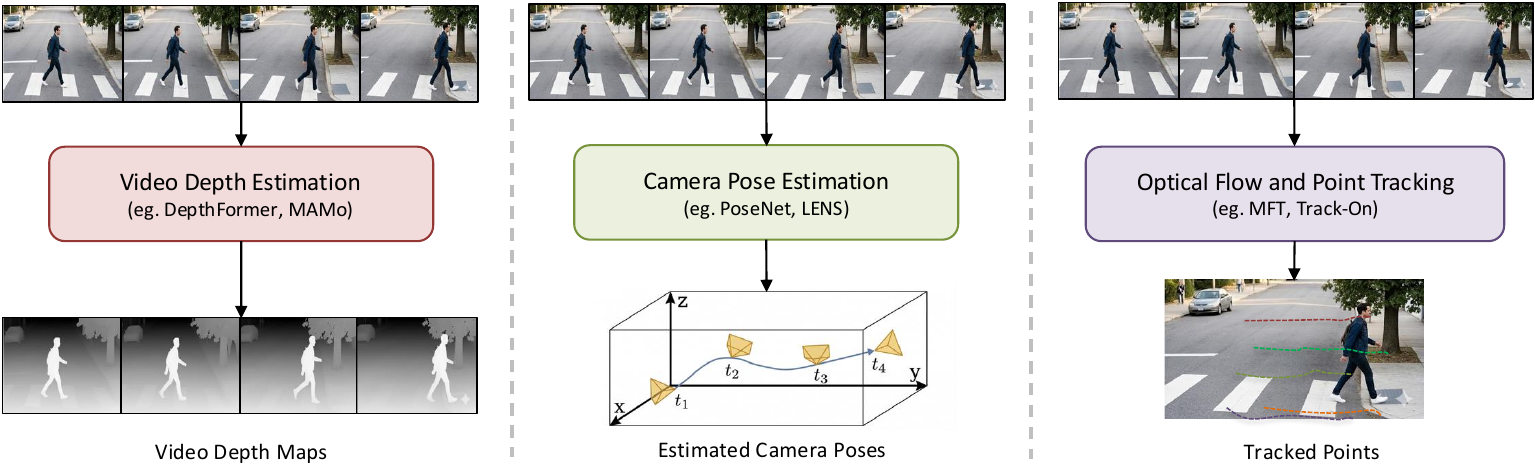}
    \caption{\textbf{Comparison of low-level video geometry understanding tasks}: video depth estimation (\textit{left}), camera pose estimation (\textit{middle}), and optical flow/point tracking (\textit{right}). While all three tasks take video frames as input, they recover different geometric quantities, namely scene depth, camera motion, and temporal correspondences across frames.}
    \label{fig:low-level-defition}
\end{figure*}

Low-level video geometry understanding studies how to recover \emph{geometry representations} directly from RGB videos. 
In contrast to high-level semantic understanding, geometry-centric tasks emphasize \emph{how the visual world is arranged and how it moves over time}.
This area has recently advanced along two complementary directions. 
First, \textbf{task-specific predictors} have become markedly more robust \emph{in the wild}, driven by foundation-scale pretraining and more scalable architecture designs. 
Second, a growing line of \textbf{joint feed-forward geometry models} moves beyond isolated tasks by predicting multiple geometric quantities within a unified network.

Accordingly, we organize this section into four parts. 
We first review video depth estimation (Section~\ref{sec:video_depth}), followed by camera pose estimation (Section~\ref{sec:camera_pose_estimation}). 
Next, we discuss optical flow and point tracking (Section~\ref{sec:point_tracking}). 
Finally, we cover joint feed-forward geometry models and their extensions (Section~\ref{sec:feedforward_geometry}).

\begin{table}[tb]
  \small
  \centering
  \renewcommand{\arraystretch}{1.1}%
  \resizebox{\linewidth}{!}{%
        \begin{tabular}{l l cc}
        \toprule
        \textbf{Method} & \textbf{Category} & Abs Rel$\downarrow$ & $\delta\uparrow$ \\
        \midrule
        DepthFormer \cite{li2023depthformer}      
            & Feed-forward        & 0.058 & \textbf{0.967} \\
        ManyDepth \cite{watson2021temporal}       
            & Feed-forward        & \textbf{0.069} & 0.930 \\
        MAMo \cite{yasarla2023mamo}                          
            & Feed-forward        & 0.049 & 0.949 \\
        NVDS \cite{NVDS}                          
            & Inference align. & 0.253 & 0.588 \\
        ChronoDepth \cite{chronodepth}    
            & Diffusion-based     & 0.167 & 0.759 \\
        Depth-Anything \cite{depth_anything_v1}   
            & Feed-forward        & 0.142 & 0.803 \\
        DepthCrafter \cite{hu2024depthcrafter}    
            & Diffusion-based     & 0.104 & 0.896 \\
        Depth Any Video \cite{yang2024depth}      
            & Diffusion-based     & 0.070 & 0.951 \\
        StableDepth \cite{zhang2025stabledepth}   
            & Diffusion-based     & 0.112 & 0.902 \\
        \bottomrule
        \end{tabular}
  }
  \vspace{4pt}
\caption{\textbf{Video depth estimation on KITTI}~\cite{geiger2013vision}\textbf{.}
Methods are grouped according to the taxonomy in Section~\ref{sec:video_depth}: inference-time alignment, feed-forward prediction, and diffusion-based approaches.
We report Absolute Relative Error (Abs Rel, $\downarrow$) and the accuracy threshold $\delta < 1.25$ ($\uparrow$).
Arrows indicate whether higher or lower values are better. Results are taken from original papers, and experimental settings may differ; strict comparisons require reference to the original protocols.}
  \label{tab:depth_kitti}
\end{table}

\subsection{Video depth estimation.}
\label{sec:video_depth}

Video depth estimation aims to predict depth maps that are both accurate and temporally coherent under viewpoint changes, occlusions, and scene dynamics, as shown in Fig.~\ref{fig:low-level-defition}. 
Compared with single-image depth estimation~\cite{liu2024depthlab,liu2024infusion}, the video setting introduces additional challenges such as temporal flicker and long-term drift, which motivate effective temporal modeling mechanisms. 
Based on this, we organize prior work into three categories: \emph{inference-time alignment}, \emph{feed-forward prediction}, and \emph{diffusion-based} approaches.
Representative quantitative results of these approaches on the KITTI benchmark are summarized in Table~\ref{tab:depth_kitti}.

\paragraph{Inference-time alignment.}
A classical strategy for achieving temporal consistency is to refine depth predictions at inference time by enforcing multi-view constraints. 
Such methods typically rely on camera pose or optical flow as motion cues and optimize per-video objectives to synchronize depth across frames~\cite{luo2020consistent,kopf2021rcvd,zhang2021consistent,chen2019self,pacheco2020inference,NVDS,NVDSPLUS}. 
While effective at reducing temporal flicker, inference-time alignment approaches often incur additional computational overhead and are sensitive to errors in the underlying motion cues, which limits their scalability to open-world and long-form videos.

\paragraph{Feed-forward prediction.}
Feed-forward methods amortize temporal reasoning into the network, predicting a sequence of depth maps directly from the input video. 
Temporal coupling is typically introduced through several design patterns. 
\emph{Matching-based} approaches aggregate multi-frame evidence using multi-view matching or cost-volume–style constructions~\cite{long2021multi,guizilini2022multi,watson2021temporal,sayed2022simplerecon,li2023depthformer}. 
\emph{Motion-guided} designs propagate or fuse features using optical flow or related motion cues~\cite{xie2020video,eom2019temporally}. 
\emph{Recurrent} formulations iteratively refine depth estimates over time~\cite{patil2020don,zhang2019exploiting}. 
Finally, \emph{attention- and memory-based} mechanisms retrieve cross-frame context to handle long-range dependencies~\cite{wang2022less,yasarla2023mamo,zhao2022monovit,yasarla2024futuredepth}. 
Other systems combine multi-frame reasoning with hybrid geometric components to balance per-frame accuracy and temporal stability~\cite{li2023temporally,depth_anything_v1}.

\paragraph{Diffusion-based approaches.}
Diffusion-based methods offer an alternative route to temporal coherence by adapting pretrained generative models for conditional depth prediction. 
By leveraging large-scale video diffusion backbones~\cite{blattmann2023stable,wan2025wan}, recent approaches synthesize depth sequences with high spatial detail and strong temporal stability across diverse scenes~\cite{chronodepth,hu2024depthcrafter,yang2024depth,zhang2025stabledepth}. 
However, these methods often inherit the substantial computational and memory costs of generative diffusion models, making long-video deployment challenging without careful windowing or streaming strategies. 
In parallel, another line of work extends strong image-generation foundations to temporally consistent video depth estimation, aiming to retain broad generalization while improving efficiency~\cite{ke2025video}.

\paragraph{Summary.}
Overall, video depth estimation has evolved from optimization-heavy inference-time alignment toward efficient feed-forward temporal modeling and diffusion-based approaches for powerful priors. 
A central open challenge remains achieving \emph{both} high geometric fidelity and long-horizon temporal stability under real-world dynamics, without incurring prohibitive computational cost.

\begin{table}[tb]
  \footnotesize
  \centering
  \renewcommand{\arraystretch}{1.1}%
  \resizebox{\linewidth}{!}{%
        \begin{tabular}{l l cc}
        \toprule
        \textbf{Method} & \textbf{Category} & $t_{\mathrm{err}}\downarrow$ & $r_{\mathrm{err}}\downarrow$ \\
        \midrule
        DenseVLAD \cite{torii201524}              
            & Correspondence & 0.26 & 13.11 \\
        Geo.\ PN \cite{kendall2017geometric}      
            & Regression (APR)              & 0.23 & 8.12  \\
        PoseNet \cite{kendall2015posenet}         
            & Regression (APR)              & 0.21 & 7.74  \\
        DFNet \cite{chen2022dfnet}                
            & Regression (APR)              & \textbf{0.02} & \textbf{0.79}  \\
        LENS \cite{moreau2022lens}                
            & Regression (APR)              & 0.08 & 3.00  \\
        ESSNet \cite{zhou2020learn}               
            & Correspondence          & 0.22 & 11.82 \\
        RelPose-GNN \cite{turkoglu2021visual}     
            & Regression (RPR)              & 0.16 & 5.20  \\
        Map-Free \cite{arnold2022map}             
            & Regression (RPR)              & 0.13 & 3.72  \\
        \bottomrule
        \end{tabular}
  }
  \vspace{4pt}
\caption{\textbf{Camera pose estimation on 7 Scenes}~\cite{shotton2013scene}\textbf{.}
Methods are grouped according to the taxonomy in Section~\ref{sec:camera_pose_estimation} into correspondence-and-solver pipelines and pose regression approaches, including absolute (APR) and relative (RPR) formulations.
We report median translation error $t_{\mathrm{err}}$ (metres, $\downarrow$) and median rotation error $r_{\mathrm{err}}$ (degrees, $\downarrow$) to reflect camera relocalisation accuracy.
Lower values indicate better accuracy. Results are taken from original papers, and experimental settings may differ; strict comparisons require reference to the original protocols.}

  \label{tab:camera}
\end{table}

\subsection{Camera pose estimation}
\label{sec:camera_pose_estimation}
As shown in Fig.~\ref{fig:low-level-defition}, camera pose estimation from videos aims to recover the 6DoF camera motion across time, and underpins localization, odometry, mapping, and 3D/4D reconstruction. 
Depending on the application, pose may be estimated (i) \emph{sequentially} from adjacent frames, (ii) from a \emph{set} of views for multi-view reconstruction, or (iii) by \emph{localizing} a query frame against a database or a pre-built map/model. 
Across these regimes, existing methods can be broadly categorized into two main families: \emph{correspondence-and-solver pipelines} and \emph{pose regression}.
Representative localization accuracy on the 7 Scenes benchmark is summarized in Table~\ref{tab:camera}.

\paragraph{Correspondence-and-solver pipelines.}
A long-standing high-precision paradigm estimates camera pose by first establishing correspondences and then applying robust geometric estimation~\cite{panek2022meshloc,yang2022scenesqueezer,giang2024learning,wang2024dgc,liu2025robust}. 
Correspondences are typically obtained through keypoint or feature matching~\cite{sun2021loftr,sarlin2020superglue,wang2024efficient,lindenberger2023lightglue,edstedt2024roma,leroy2024grounding}, or via scene-coordinate regression that directly predicts pixel-to-3D associations~\cite{brachmann2021visual,tang2021learning,dong2022visual,brachmann2023accelerated,wang2024hscnet++}. 
Given noisy matches, robust estimators such as RANSAC variants combined with minimal solvers (\eg, 5/7/8-point) recover relative pose up to scale, or absolute pose when metric information is available~\cite{fischler1981random,barath2019magsac,barath2020magsac++}.  

This paradigm also underlies many sequential pose estimation pipelines~\cite{zhou2022edplvo,kannapiran2023stereo,shu2023structure,jiang2024ul,zhou2020learn}, where pose is estimated incrementally between adjacent frames using tracked features or direct photometric alignment, yielding temporally consistent trajectories. 
Due to their strong geometric guarantees, correspondence-based approaches form the backbone of many practical localization and reconstruction systems~\cite{li2010location,sattler2017large,taira2018inloc,sattler2018benchmarking,sarlin2021back,torii201524}. 
However, their runtime is often dominated by matching and robust estimation, and scaling supervision can be constrained by the need for explicit correspondences or map-specific signals.

\paragraph{Pose regression: absolute and relative.}
An alternative to correspondence-heavy pipelines is end-to-end pose regression, valued for its simplicity and real-time inference~\cite{kendall2015posenet,kendall2017geometric}. 
Absolute pose regression (APR) predicts camera pose directly in a world coordinate system; while efficient, APR often trails correspondence-based methods in accuracy and may behave similarly to retrieval~\cite{sattler2019understanding} unless trained with dense viewpoint coverage or strong scene priors~\cite{sattler2019understanding,moreau2022lens,chen2022dfnet,chen2024neural,lin2024learning}. 
Relative pose regression (RPR)~\cite{abouelnaga2021distillpose,khatib2022leveraging,arnold2022map,turkoglu2021visual,zhang2022relpose,wang2023posediffusion,lin2024relpose++,zhang2024cameras,tu2024panopose} instead predicts the relative transformation between image pairs or short temporal windows.
Compared to APR, RPR often generalizes better across scenes and can be naturally applied to sequential video data by composing relative estimates over time~\cite{shen2023dytanvo,teed2024deep,xu2025airslam,lipson2024deep,wimbauer2025anycam,rockwell2025dynamic}. 
Moreover, relative predictions can be upgraded to absolute localization by aggregating multiple retrieved pairs or through multi-view optimization~\cite{laskar2017camera,winkelbauer2021learning}.  
Recent efforts emphasize \emph{scaling} relative pose regression to large and diverse datasets to improve robustness across object-centric, indoor, and outdoor domains, exemplified by large-scale training pipelines~\cite{rockwell2024far,dong2025reloc3r}.

\paragraph{Summary.}
Overall, correspondence-and-solver pipelines remain a high-precision, geometry-grounded standard, while pose regression offers a lightweight and increasingly competitive alternative as training data and model capacity scale. 
In later sections, we revisit pose estimation in the context of \emph{joint} feed-forward geometry models, where pose is predicted together with other geometric primitives, highlighting the benefits of learning in conjunction with complementary modalities.

\begin{table}[tb]
  \footnotesize
  \centering
  \renewcommand{\arraystretch}{1.1}%
  \resizebox{\linewidth}{!}{%
        \begin{tabular}{l l ccc}
        \toprule
        \textbf{Method} & \textbf{Category} & AJ$\uparrow$ & $\delta_{\mathrm{avg}}\uparrow$ & OA$\uparrow$ \\
        \midrule
        MFT \cite{mft}                            
            & Dense flow-based      & 47.3 & 66.8 & 77.8 \\
        TAPIR \cite{doersch2023tapir}                        
            & Offline TAP          & 56.2 & 70.0 & 86.5 \\
        SpatialTracker \cite{xiao2024spatialtracker} 
            & 3D-consistent         & 61.1 & 76.3 & 89.5 \\
        TAPTRv2 \cite{li2024taptrv2}           
            & Offline TAP          & 63.5 & 75.9 & 91.4 \\
        CoTracker3 \cite{karaev2024cotracker3}    
            & Offline TAP     & 64.5 & 76.7 & 89.7 \\
        Track-On \cite{aydemir2025track}                  
            & Online TAP         & \textbf{65.0} & \textbf{78.0} & \textbf{90.8} \\
        \bottomrule
        \end{tabular}
  }
  \vspace{4pt}
\caption{\textbf{Optical flow and point tracking on DAVIS}~\cite{pont20172017}\textbf{.}
Methods are categorized according to Section~\ref{sec:point_tracking} into dense flow-based tracking, offline Tracking Any Point (TAP), and online TAP.
Tracking quality is evaluated using Average Jaccard (AJ, $\uparrow$), average displacement accuracy $\delta_{\mathrm{avg}}$ ($\uparrow$), and overall accuracy (OA, $\uparrow$).
Higher values indicate better performance. Results are taken from original papers, and experimental settings may differ; strict comparisons require reference to the original protocols.} 
  \label{tab:low_tracking}
\end{table}

\subsection{Optical flow and point tracking}
\label{sec:point_tracking}
Point tracking provides a fundamental notion of \emph{temporal correspondence}: given a point in one frame, the goal is to recover its trajectory through time despite viewpoint changes, occlusions, and appearance variation, as illustrated in Fig.~\ref{fig:low-level-defition}. This capability is central to motion analysis, dynamic reconstruction, and consistency-aware generation, and it is closely related to optical flow.
Representative performance of recent tracking methods on the DAVIS benchmark is summarized in Table~\ref{tab:low_tracking}.

\paragraph{From short-range optical flow to long-range correspondence.}
Optical flow estimates dense, frame-to-frame displacement fields and has a long history, from classical formulations such as Lucas--Kanade~\cite{baker2004lucas} to modern deep networks including FlowNet~\cite{dosovitskiy2015flownet}, RAFT~\cite{teed2020raft}, and dense correspondence trackers such as MFT~\cite{mft}.
While highly effective for capturing local motion between adjacent frames, optical flow does not explicitly maintain point identities over time; chaining flow across many frames can accumulate drift and is brittle under long occlusions or re-appearance.

Long-term point tracking addresses these limitations by explicitly modeling correspondence \emph{across time}. 
This problem is often framed as \emph{Tracking Any Point (TAP)}~\cite{tapvid}, which seeks to recover trajectories for arbitrary points over extended horizons, including through occlusion and re-entry. 
Many TAP methods reuse flow-inspired components for local matching, but augment them with longer temporal context, explicit visibility reasoning, and memory mechanisms to preserve correspondence over time~\cite{doersch2023tapir,doersch2024bootstap}.

\begin{figure*}[!ht]
    \setlength{\abovecaptionskip}{3mm}
    \centering
    \includegraphics[width=0.999\textwidth]{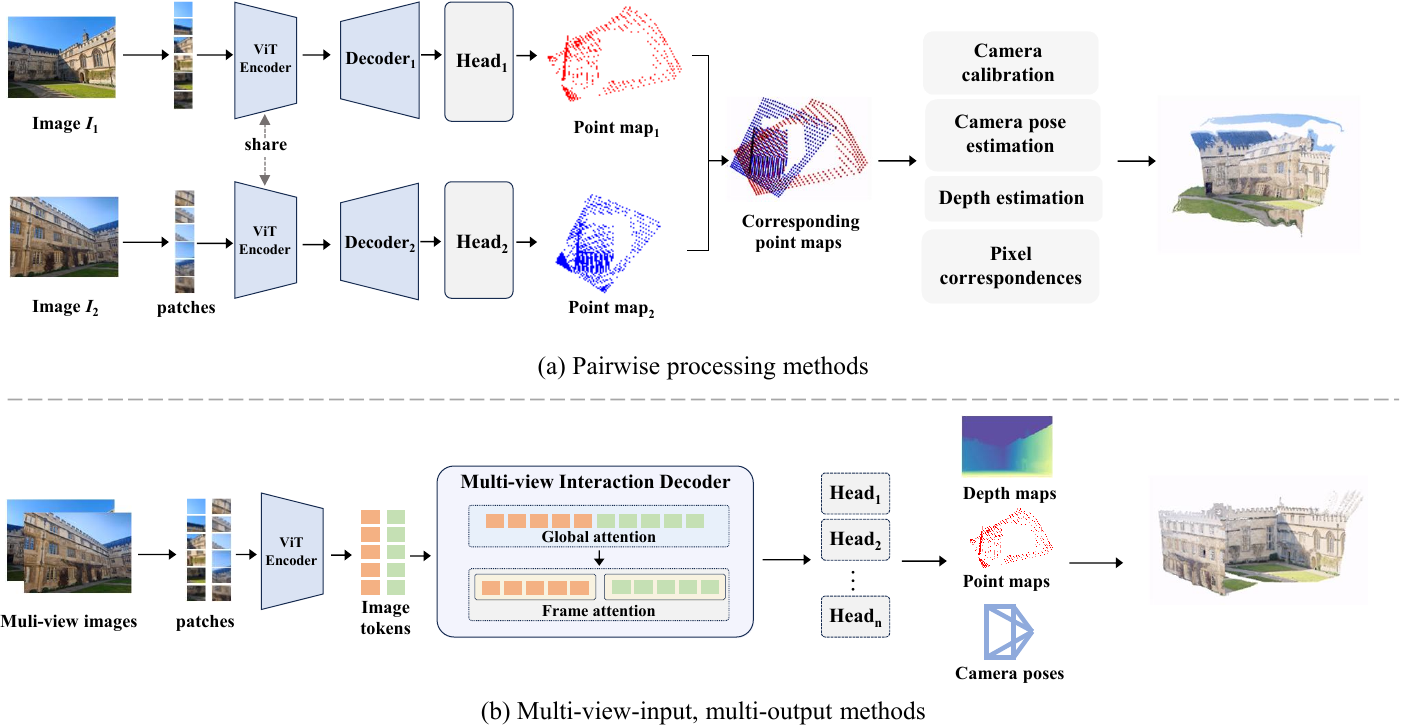}
\caption{\textbf{Comparison between different joint feed-forward geometry models.}
(a) \emph{Pairwise processing} methods~\cite{weinzaepfel2022croco,wang2024dust3r} predict view-consistent representations (\eg, point maps) from image pairs, typically operating on independent frame pairs and relying on later aggregation or optimization to get multi-view geometry information. 
(b) \emph{Multi-view-input, multi-output} methods~\cite{wang2025vggt,cut3r} jointly reason over a set of frames in a single forward pass, using global attention or memory to directly predict multiple geometric primitives with cross-view consistency.
}
\label{fig:section2}
\end{figure*}

\paragraph{Progress in Tracking Any Point.}
Most TAP methods follow a common template: (i) compute a correspondence signal typically via feature matching and correlation, and (ii) iteratively refine trajectories over time~\cite{sand2008particle,harley2022particle}. 
Subsequent work~\cite{tapvid,doersch2023tapir,li2024taptr,li2024taptrv2,qu2024taptrv3,doersch2024bootstap,kim2025learning,balasingam2024drivetrack,jin2024stereo4d} has expanded this framework along several directions.
Some methods explore stronger global reasoning, including joint multi-point tracking that exploits correlations among trajectories~\cite{karaev2024cotracker,karaev2024cotracker3}, region-level matching with enlarged receptive fields~\cite{cho2024local}, motion-guided refinement~\cite{cho2024flowtrack,le2024dense}, and test-time optimization for challenging motion and occlusion patterns~\cite{tumanyan2024dino,wang2023tracking}.  
Other works demonstrate that stronger and more diverse feature representations can substantially improve correspondence quality: foundation models such as DINOv2~\cite{oquab2023dinov2} yield consistent gains~\cite{aydemir2024can,kim2025exploring,aydemir2025track}, while diffusion-based representations further broaden the design space~\cite{nam2025emergent,son2025repurposing}. 
Finally, there is growing interest in producing \emph{3D-consistent} trajectories by coupling tracking with depth and camera motion~\cite{song2024track,xiao2024spatialtracker,wang2024scenetracker,ngo2024delta,cho2025seurat,zhang2025tapip3d} or predicting a \emph{trajectory field} that assigns every pixel a continuous 3D trajectory~\cite{liu2025trace}.

\begin{table*}[t]
\centering
\setlength{\tabcolsep}{4pt}
\renewcommand{\arraystretch}{1.1}%
\renewcommand{\arraystretch}{0.95}
\footnotesize
\begin{tabular}{l c cc cc cc cc cc}
\toprule
\multicolumn{1}{c}{\textbf{Method}} &
\multicolumn{1}{c}{\textbf{Type}} &
\multicolumn{4}{c}{\textbf{Video Depth (VD)}} &
\multicolumn{2}{c}{\textbf{Camera Pose}} &
\multicolumn{4}{c}{\textbf{3D Reconstruction}} \\
\cmidrule(lr){3-6} \cmidrule(lr){7-8} \cmidrule(lr){9-12}

&
& \multicolumn{2}{c}{KITTI}
& \multicolumn{2}{c}{Sintel}
& \multicolumn{1}{c}{7 Scenes}
& \multicolumn{1}{c}{RSE10K}
& \multicolumn{2}{c}{Co3D v2}
& \multicolumn{2}{c}{NRGBD} \\
\cmidrule(lr){3-4} \cmidrule(lr){5-6}
\cmidrule(lr){7-8}  \cmidrule(lr){9-10} \cmidrule(lr){11-12}

&
& Abs Rel$\downarrow$ & $\delta\uparrow$
& Abs Rel$\downarrow$ & $\delta\uparrow$
& A30$\uparrow$ & A30$\uparrow$
& Acc.$\downarrow$ & Comp.$\downarrow$
& Acc.$\downarrow$ & Comp.$\downarrow$ \\
\midrule

DUST3R \cite{wang2024dust3r}      & Pair \& MV 
& 0.143 & 0.814 & 0.653 & 44.9 & 76.7 & 67.7 & 0.077 & 0.067 & 0.019 & 0.018 \\

MASt3R \cite{leroy2024grounding}     & Pair \& MV 
& 0.115 & 0.848 & 0.641 & 43.9 & 81.8 & 76.4 & 0.081 & 0.069 & 0.033 & 0.028 \\

MONST3R \cite{zhang2024monst3r}   & Dynamic 
& 0.107 & 0.884 & 0.358 & 54.8 & -- & -- & 0.185 & 0.167 & 0.114 & 0.110 \\

Spann3R \cite{wang2024spann3r}                         & Dynamic
& 0.198 & 0.837 & 0.622 & 42.6 & 60.8 & 70.4 & 0.226 & 0.112 & 0.323 & 0.285 \\

Fast3R \cite{Yang_2025_fast3R}      & Dynamic 
& 0.138 & 0.834 & 0.653 & 44.9 & 75.0 & 72.7 & \textbf{0.016} & \textbf{0.009} & 0.034 & \textbf{0.010} \\

MV-DUST3R \cite{tang2024mv-dust3r+} & Pair \& MV   
& 0.456 & 0.342 & -- & -- & 69.5 & 71.3 & -- & -- & -- & -- \\

CUT3R \cite{cut3r}                & Dynamic 
& 0.122 & 0.876 & 0.421 & 42.4 & 82.8 & 75.3 & 0.047 & 0.031 & 0.031 & 0.026 \\

FLARE \cite{zhang2025flare}       & Pair \& MV 
& 0.356 & 0.570 & -- & -- & 83.3 & 78.8 & 0.057 & 0.107 & 0.024 & 0.025 \\

VGGT \cite{wang2025vggt}    & Pair \& MV 
& \textbf{0.062} & \textbf{0.969} & \textbf{0.297} & \textbf{68.8} & 87.7 & 85.3 & 0.039 & 0.039 & 0.018 & 0.021 \\

$\pi$3 \cite{wang2025pi}  & Dynamic 
& 0.038 & 0.986 & -- & -- 
& \textbf{88.4} & 85.9
& 0.029 & 0.049 & \textbf{0.015} & 0.014 \\

StreamVGGT \cite{zhuo2025streamVGGT}                       & Dynamic
& 0.072 & 94.7 & 0.323 & 65.7 & 82.4 & -- & 0.056 & 0.041 & 0.044 & 0.041 \\

\bottomrule
\end{tabular}
\vspace{4pt}
\caption{
\textbf{Comparison of feed-forward geometry models across video depth, camera pose, and 3D reconstruction tasks.}
\textbf{VD} denotes \emph{Video Depth Estimation}.
For VD benchmarks (\textbf{KITTI} and \textbf{Sintel}), performance is evaluated using \emph{Absolute Relative Error} (Abs Rel, $\downarrow$) and the depth accuracy metric $\delta$ ($\uparrow$), where $\delta$ denotes the percentage of pixels satisfying $\max(\hat{d}/d, d/\hat{d}) < 1.25$.
\textbf{Camera pose estimation} is evaluated on \textbf{7 Scenes} and \textbf{RealEstate10K} using \emph{A30} ($\uparrow$), defined as the area under the error curve up to $30^\circ$ for translation and rotation errors.
\textbf{3D reconstruction} performance is reported on \textbf{Co3D v2} and \textbf{NRGBD} using reconstruction \emph{accuracy} (Acc., $\downarrow$) and \emph{completeness} (Comp., $\downarrow$).
Arrows indicate whether higher or lower values are better. ``Pair \& MV” denotes methods based on pairwise or multi-view paradigms, while ``Dynamic” refers to approaches designed for dynamic scenes and online deployment. Results are taken from original papers, and experimental settings may differ; strict comparisons require reference to the original protocols.
}
\label{tab:joint-ffg}
\end{table*}

\paragraph{Online tracking for long-horizon videos.}
Most high-performing TAP models operate offline or within fixed temporal windows, allowing access to future frames during inference~\cite{karaev2024cotracker,karaev2024cotracker3}. 
In contrast, streaming scenarios require \emph{online} prediction, where future frames are unavailable, and occlusions may span long durations. 
This motivates online trackers that maintain explicit memory or recurrent state for long-horizon reasoning~\cite{aydemir2025track,vecerik2024robotap,patraucean2023perception,aydemir2025track,zholus2025tapnext}. 
Online tracking is particularly important for robotics and interactive systems with strict latency constraints~\cite{abbeel,vecerik2024robotap}, and is increasingly used as a control signal in generation and editing pipelines that demand temporally consistent guidance~\cite{zhou2024trackgo}.

\paragraph{Summary.}
In summary, optical flow provides a strong foundation for short-range motion estimation, while TAP formulations extend correspondence to long-term, identity-preserving trajectories under occlusion. 
Progress is driven by advances in global reasoning, multi-point interaction, and increasingly strong feature representations from foundation and diffusion models. 
A key frontier remains \emph{streaming, long-horizon} tracking that is robust to extended occlusions and dynamic scenes while remaining efficient enough for real-time use.

\subsection{Joint feed-forward geometry models}
\label{sec:feedforward_geometry}

After reviewing task-specific predictors, we now turn to a rapidly growing family of \emph{joint} feed-forward geometry models, which aim to predict multiple geometric primitives (\eg, camera pose, depth maps, point maps) within a single network pass. 
These models are particularly attractive as they enable fast inference without heavy test-time optimization, while providing a unified geometric interface for video understanding and reconstruction~\cite{smart2024splatt3r,fei2024driv3r,dong2025reloc3r,leroy2024grounding,fang2025dens3r,wang2025pi}.
Table~\ref{tab:joint-ffg} summarizes the performance of various methods across multiple tasks and datasets, while Fig.~\ref{fig:section2} provides a schematic comparison of representative joint feed-forward geometry architectures. 
As illustrated in Fig.~\ref{fig:section2}, joint feed-forward geometry models aim to recover a coherent 3D scene understanding by jointly solving multiple geometric subproblems in a single forward pass, yielding mutually consistent geometric representations. These typically include camera calibration, camera pose estimation, depth estimation, and pixel correspondence prediction. 
Camera calibration estimates intrinsic parameters that map pixels to camera rays, while camera pose estimation recovers the 6-DoF extrinsic transformation, parameterized by rotation $R$ and translation $t$, that defines the camera position and orientation in a reference frame. 
Depth estimation predicts per-pixel distance along the viewing direction, and pixel correspondences identify the same physical 3D point across frames. Together, these components provide the foundation for multi-view consistency and geometry fusion across time and viewpoints.

\paragraph{From pairwise point maps to multi-view, multi-output prediction.}
DUSt3R~\cite{wang2024dust3r} introduced the core idea of regressing view-consistent point maps from an image pair without requiring camera calibration at inference time. 
A key development for video is moving beyond pairwise processing: recent models incorporate multi-frame context via memory encoders and factorized or subgraph-based fusion~\cite{wang2024spann3r,cabon2025must3r,liu2024slam3r}, or by adopting global-attention multi-view transformers that jointly reason over a set of frames in a single forward pass~\cite{Yang_2025_fast3R,wang2025vggt,lin2025depth}. 

In parallel, these models increasingly exhibit \emph{jointness} in their outputs through multi-task prediction. 
Beyond point maps, they simultaneously predict camera poses, dense depth, and tracking-related features within a unified model~\cite{wang2025vggt,cut3r}, indicating that learning shared representations across multiple geometric primitives can improve both performance and robustness. 
Some approaches further connect these feed-forward priors to 3D reconstruction, enabling end-to-end pipelines that directly optimize or render a persistent 3D representation~\cite{tang2024mv-dust3r+,zhang2025flare,kerbl2023-3dgs}. Building on these persistent representations, recent advancements leverage continuous modeling techniques, such as Gaussian Splatting and dense point cloud regression, to bridge scene reconstruction with controllable 3D scene editing and panorama generation~\cite{zhang2026psgs,chen2025dustgs,oehmcke2024lidar}. Furthermore, integrating these geometric foundations with controllable frameworks allows for highly customized, text-driven 3D scene synthesis~\cite{liu2024graph}.

\paragraph{Dynamic scenes and online deployment.}
Applying joint feed-forward geometry models to real videos introduces two requirements that are less critical in static multi-view settings: handling \emph{scene dynamics} and supporting \emph{incremental} updates as new frames arrive. 
Historically, dynamic reconstruction relied on motion or semantic masking combined with per-sequence optimization, often powered by strong depth and flow priors~\cite{qiu2022airdos,henein2020dynamic,li2024megasam,wu2024cat4d,kopf2021rcvd,lei2025mosca,matsuki20254dtam}. 
To more effectively handle dynamics, recent works adapt feed-forward geometry priors to dynamic scenes through fine-tuning with motion cues, auxiliary motion-mask heads, inference-time attention adaptation, or factorized prediction schemes~\cite{zhang2024monst3r,xu2024das3r,chen2025easi3r,karhade2025any4d,hu2025vggt4d}. 

Meanwhile, many multi-view transformers remain inherently \emph{offline}: set-based inference and global attention often require re-encoding the full frame set when new observations arrive~\cite{wang2025vggt}. 
This limitation motivates streaming formulations that maintain an explicit scene state for incremental updates, including spatial or pointer-based memory designs~\cite{wang2024spann3r,wu2025point3r}, recurrent read–write scene representations~\cite{cut3r}, and causal streaming variants designed for long sequences~\cite{lan2025stream3r,zhuo2025streamVGGT}. 
Together, these directions push joint feed-forward geometry toward practical video understanding in dynamic, long-horizon, and low-latency settings.

\paragraph{Summary.}

Overall, the feed-forward geometry models have evolved from pairwise point-map regression~\cite{wang2024dust3r} to multi-view architectures that jointly predict multiple geometric outputs~\cite{cut3r,wang2025vggt}, and further toward dynamic and streaming deployment through motion-aware adaptation and memory for online inference~\cite{wang2024spann3r,wu2025point3r,lan2025stream3r,zhuo2025streamVGGT}. 
This line of work forms a direct bridge from task-specific geometric estimation to unified geometry understanding, producing compact geometric primitives that can be readily consumed by higher-level reasoning and generation modules.

\begin{figure*}[!t]
    \setlength{\abovecaptionskip}{3mm}
    \centering
    \includegraphics[width=0.999\linewidth]{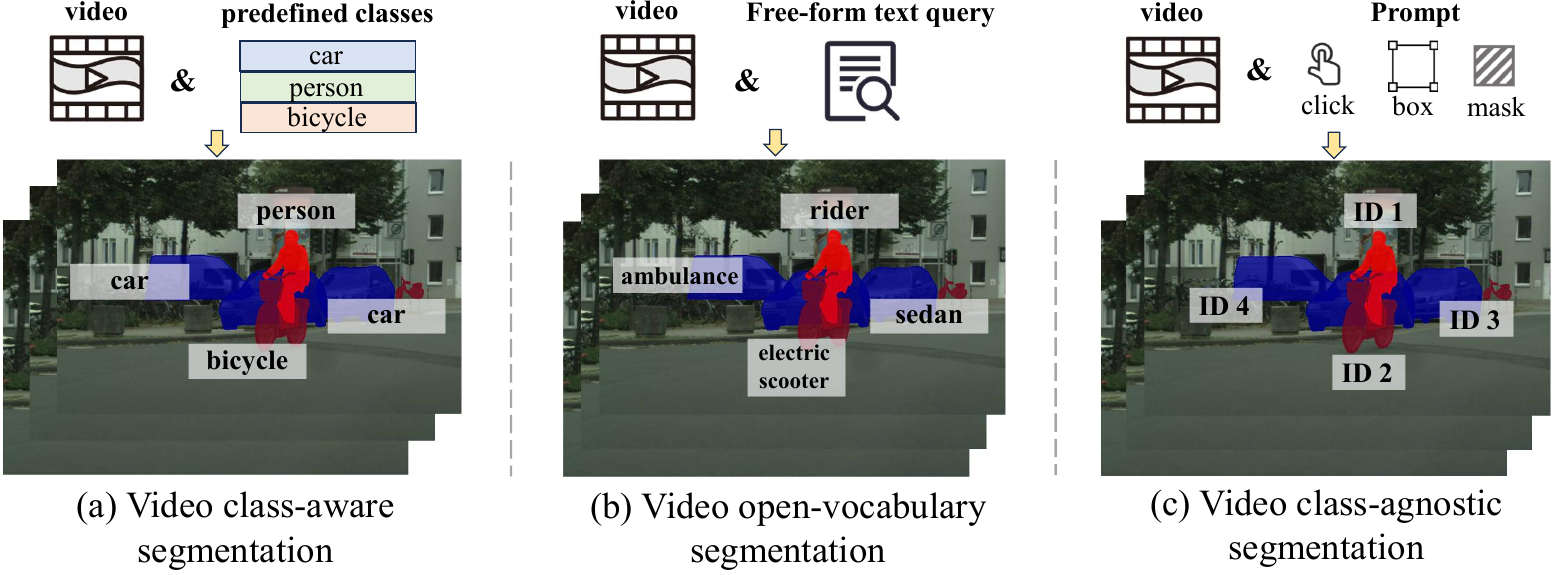}
\caption{\textbf{Overview of video segmentation.}
We categorize video segmentation methods into three paradigms based on how semantic categories are specified.
(a) \emph{Video class-aware segmentation} assumes a fixed, predefined label set and includes video semantic, instance, and panoptic segmentation.
(b) \emph{Video open-vocabulary segmentation} extends class-aware settings by leveraging language embeddings to segment both seen and unseen categories.
(c) \emph{Video class-agnostic segmentation} removes semantic labels altogether, instead segmenting and tracking objects using visual or concept-level prompts such as clicks, boxes, or masks.}
\label{fig:segmentation}
\end{figure*}

\section{High-level Video Semantic Understanding}
\label{sec:high_level_semantics}

While low-level geometry recovers physically grounded structure, high-level semantic understanding~\cite{an2025generalized,an2024multimodality,thengane2026scope,an2024rethinking,li2025chatmotion} seeks to infer \emph{what the world means}: what entities, actions, and events are present in a video, and how they evolve over time. 
Compared with image-level analysis~\cite{wu2023object,sun2023indiscernible,gui2024knn,ren2023masked,ren2024sharing}, video semantics places stronger demands on \textbf{spatiotemporal reasoning}, including handling motion, occlusion, and long-range temporal dependencies, as well as on \textbf{cross-modal grounding}, which aligns visual evidence with language queries. 
As a result, high-level video understanding requires models to integrate information across time, maintain consistent semantic representations under various changes, and localize the visual evidence that supports semantic decisions.

Motivated by these requirements, we focus this section on video segmentation, object tracking, and temporal grounding, as they emphasize explicit spatiotemporal localization, identity-preserving modeling, and compatibility with multimodal foundation models. 
Together, these tasks capture complementary aspects of high-level semantics, including spatial understanding, identity consistency, and language-aligned temporal reasoning. 
Beyond our main scope, classical high-level video semantics has also been extensively studied through tasks such as action recognition~\cite{peng2024referring,yang2024adapting,han2025training,wang2025foundation} and video captioning~\cite{meng2025videocap,lin2022swinbert}, which typically assign clip-level labels or generate free-form textual summaries of events. 
These tasks are also important and have driven progress in video semantic learning.
With this scope in mind, we organize this section around three representative tasks:
(i) \textbf{video segmentation} (Section~\ref{sec:video_segmentation}), 
(ii) \textbf{video object tracking} (Section~\ref{sec:video_object_tracking}), and 
(iii) \textbf{video temporal grounding} (Section~\ref{sec:video_grounding}).

\subsection{Video segmentation}
\label{sec:video_segmentation}
Video segmentation aims to assign a label or mask to each pixel according to specific properties or semantics and to maintain temporal consistency over time~\cite{zhou2022survey}. Depending on whether the segmentation categories are known, we divide it into three types of tasks, as illustrated in Fig.~\ref{fig:segmentation}.

\paragraph{Video class-aware segmentation.}
Video class-aware segmentation methods can be categorized by segmentation objective into video semantic segmentation (VSS), video instance segmentation (VIS), and video panoptic segmentation (VPS). VSS aims to extract objects belonging to predefined semantic categories (\eg, cars, pedestrians, roads) from videos. 
A major line of VSS improves cross-frame accuracy through optical-flow feature warping and aggregation~\cite{jin2017video,huang2018efficient,nilsson2018semantic}, spatio-temporal interaction~\cite{hesham2025exploiting,weng2023mask,sun2022coarse,sun2022mining,an2023temporal,ariff2025evaluating}, or diffusion-based methods~\cite{wang2025vidseg,delatolas2025studying} to enforce temporal consistency and leverage multi-frame context. 
Another complementary line of work leverages temporal information to speed up video segmentation by avoiding full per-frame computation~\cite{yang2024end,hu2023efficient,lin2024exploring}, typically via feature reuse across neighboring frames~\cite{weng2023mask}. 
VIS requires simultaneous detection, segmentation, and tracking of instances in videos~\cite{yang2019video}. 
While some methods~\cite{liu2021sg,wu2021track,fu2021compfeat,wu2022defense,zhou2024improving} detect and segment instances for each individual frame, followed by frame-by-frame instance tracking, an alternative paradigm is to segment-as-a-whole by posing the task as a direct sequence prediction problem~\cite{li2023tcovis,zhang2023dvis,ying2023ctvis,dong2025hierarchical,lee2025cavis}. 
VPS associates object instances across frames to form temporally consistent tracklets, while simultaneously performing fine-grained pixel-level panoptic labeling in each frame.
This is typically achieved through coarse segment-level matching for instance association and fine-grained pixel-level segmentation across time~\cite{kim2020video,woo2021learning,shin2024video,choudhuri2023context}. 

\begin{figure*}[!t]
    \setlength{\abovecaptionskip}{3mm}
    \centering
    \includegraphics[width=0.999\linewidth]{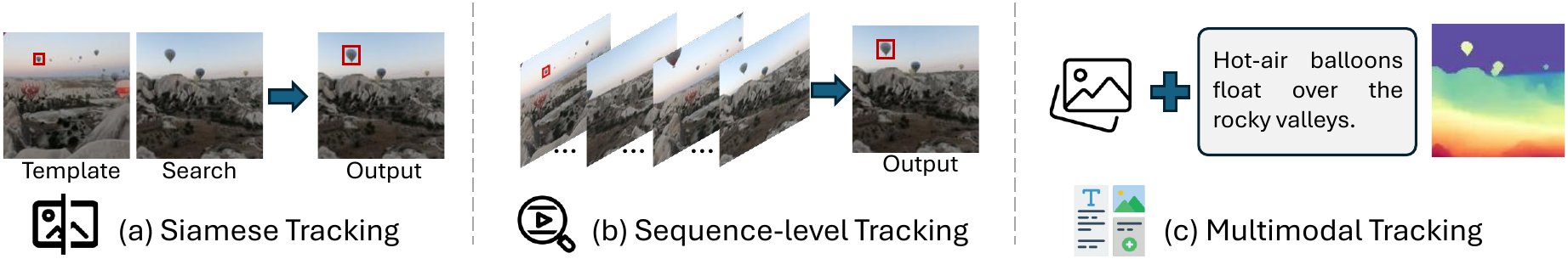}
\caption{\textbf{Overview of video object tracking.} (a) Siamese tracking performs target localization by matching a template extracted from the target object against a search region in the current frame. (b) Sequence-level tracking extends this formulation by modeling temporal dependencies across multiple consecutive frames, thereby improving tracking continuity and robustness. (c) Multimodal tracking further incorporates complementary modalities, such as visual, textual, or auxiliary sensory cues, to enhance target representation and achieve more reliable tracking in challenging scenarios.}
\label{fig:tracking}
\end{figure*}

\paragraph{Video open-vocabulary segmentation.}
Video open-vocabulary segmentation methods segment objects in a video using free-form text categories, including unseen classes, aligning pixel-level masks with open-ended textual concepts. 
Powerful pretrained vision-language models (VLMs)~\cite{radford2021learning,tschannen2025siglip} enable impressive zero-shot object recognition capabilities in tasks such as classification, captioning, and segmentation. 
Wang et al.~\cite{wang2023towards} first introduce open-vocabulary VIS and achieve end-to-end inference with an efficient memory-induced Transformer architecture. 
Subsequently, memory-based methods~\cite{zhu2024clip,wang2024ov,guo2025openvis} further strengthen cross-frame instance matching, while later works~\cite{zhang2025dvis++,fang2024unified} explore decoupled adapters to better predict mask proposals and attention bias.
In parallel, another line of work extends open-vocabulary capabilities to pixel-level semantic understanding, giving rise to open-vocabulary video semantic segmentation (VSS). 
Recent studies~\cite{li2025towards,xu2023side,zhu2025sed++,shin2024towards} primarily focus on transferring open-set linguistic knowledge from pretrained vision–language models to dense prediction tasks by aligning pixel-level representations with text embeddings, enabling segmentation beyond predefined category sets.

\begin{table*}[t]
\centering
\renewcommand{\arraystretch}{1.1}
\resizebox{\textwidth}{!}{
\begin{tabular}{l c c c c c c c c c c c}
\toprule
\multirow{2}{*}{\textbf{Method}} & \multirow{2}{*}{\textbf{Modality}} &
\multicolumn{3}{c}{\textbf{DepthTrack}~\cite{yan2021depthtrack}} &
\multicolumn{2}{c}{\textbf{LasHeR}~\cite{li2021lasher}} &
\multicolumn{2}{c}{\textbf{RGBT234}~\cite{li2019rgb}} &
\multicolumn{2}{c}{\textbf{VisEvent}~\cite{wang2023visevent}} \\
\cmidrule(lr){3-5}\cmidrule(lr){6-7}\cmidrule(lr){8-9}\cmidrule(lr){10-11}
& & \textbf{F-score} $\uparrow$ & \textbf{Recall} $\uparrow$ & \textbf{Precision} $\uparrow$ &
\textbf{Precision} $\uparrow$ & \textbf{AUC} $\uparrow$ &
\textbf{MPR} $\uparrow$ & \textbf{MSR} $\uparrow$ &
\textbf{Precision} $\uparrow$ & \textbf{AUC} $\uparrow$ \\
\midrule

\multirow{2}{*}{ViPT~\cite{zhu2023visual}}
& Miss & 44.4 & 40.5 & 46.6 & 40.1 & 34.0 & 52.4 & 39.4 & 57.2 & 43.2 \\
& Full & 59.4\gain{15.0} & 59.6\gain{19.1} & 59.2\gain{12.6} & 65.1\gain{25.0} & 52.5\gain{18.5} & 83.5\gain{31.1} & 61.7\gain{22.3} & 75.8\gain{18.6} & 59.2\gain{16.0} \\
\midrule

\multirow{2}{*}{SDSTrack~\cite{hou2024sdstrack}}
& Miss & 46.7 & 42.0 & 52.7 & 52.5 & 43.1 & 67.0 & 48.8 & 62.6 & 46.9 \\
& Full & 61.4\gain{14.7} & 60.9\gain{18.9} & 61.9\gain{9.2} & 66.5\gain{14.0} & 53.1\gain{10.0} & 84.8\gain{17.8} & 62.5\gain{13.7} & 76.7\gain{14.1} & 59.7\gain{12.8} \\
\midrule

\multirow{2}{*}{STTrack~\cite{hu2025exploiting}}
& Miss & 49.9 & 48.8 & 51.0 & 54.5 & 44.9 & 73.8 & 54.2 & 65.5 & 49.7 \\
& Full & 63.3\gain{13.4} & 63.4\gain{14.6} & 63.2\gain{12.2} & 76.0\gain{21.5} & 60.3\gain{15.4} & 89.8\gain{16.0} & 66.7\gain{12.5} & 78.6\gain{13.1} & 61.9\gain{12.2} \\
\midrule

\multirow{2}{*}{SUTrack~\cite{chen2024sutrack}}
& Miss & 49.5 & 47.3 & 51.9 & 58.3 & 47.6 & 82.0 & 60.8 & 66.6 & 50.5 \\
& Full & 65.1\gain{15.6} & 65.7\gain{18.4} & 64.5\gain{12.6} & 74.5\gain{16.2} & 59.5\gain{11.9} & 92.2\gain{10.2} & 69.5\gain{8.7} & 79.9\gain{13.3} & 62.7\gain{12.2} \\
\midrule

\multirow{2}{*}{FlexTrack~\cite{tan2025you}}
& Miss & 57.8 & 56.1 & 59.6 & 65.1 & 52.3 & 84.1 & 62.6 & 72.8 & 55.0 \\
& Full & \textbf{67.0}\gain{9.2} & \textbf{66.9}\gain{10.8} & \textbf{67.1}\gain{7.5} & \textbf{77.3}\gain{12.2} & \textbf{62.0}\gain{9.7} & \textbf{92.7}\gain{8.6} & \textbf{69.9}\gain{7.3} & \textbf{81.4}\gain{8.6} & \textbf{64.1}\gain{9.1} \\

\bottomrule
\end{tabular}
}
\vspace{5pt}
\caption{\textbf{Performance comparison on four multimodal tracking benchmarks} under \textbf{Full} (complete modalities) and \textbf{Miss} (temporally missing modalities) settings, as introduced in \cite{tan2025you}. Missing-setting results are reported on the corresponding \texttt{*\textsubscript{miss}} benchmark variants including \texttt{DepthTrack\textsubscript{miss}}, \texttt{LasHeR\textsubscript{miss}}, \texttt{RGBT234\textsubscript{miss}}, and \texttt{VisEvent\textsubscript{miss}}. Performance gaps between Full and Miss settings are highlighted in \textcolor{teal}{green}; smaller gaps indicate stronger robustness to modality absence. Results are taken from original papers, and experimental settings may differ; strict comparisons require reference to the original protocols.}
\label{tab:tracking}
\end{table*}

\paragraph{Video class-agnostic segmentation.}

Video class-agnostic segmentation aims to segment and track object instances without predefined semantic labels, relying instead on user-provided visual prompts such as clicks or bounding boxes.
Early approaches~\cite{cheng2023tracking,siam2021video} typically adopt temporal propagation frameworks, in which masks predicted on a single frame are propagated across the entire video.
Motivated by scaling laws observed in large language models, SAM2~\cite{ravi2024sam2} introduces a simple yet scalable transformer architecture trained on large-scale data, achieving strong zero-shot performance for video class-agnostic segmentation. 
Its success has spurred a wide range of downstream applications, such as medical video segmentation~\cite{zhu2024medical,ji2022video},long-video segmentation and tracking~\cite{ding2025sam2long}, and embodied AI~\cite{deng2025geosam2}. 
More recently, SAM3~\cite{carion2025sam} expands upon SAM2 by incorporating higher-level concept prompts, such as short textual phrases and exemplar images, enabling segmentation in both class-agnostic and open-vocabulary scenarios. 
Architecturally, SAM3 moves beyond SAM2’s propagation-centric design by unifying concept-conditioned detection and memory-based video tracking within a shared-backbone framework, allowing seamless integration of concept-level prompts.

\paragraph{Summary}
Video segmentation studies the problem of assigning a label or mask to every pixel in a video while keeping the predictions consistent over time. Early research largely focused on task-specific designs to exploit temporal cues for stability and accuracy across frames, then gradually shifted toward more scalable and unified formulations that jointly handle object segmentation and tracking in videos. More recently, progress in vision-language and segmentation foundation models has expanded video segmentation from closed-set settings to open-vocabulary and prompt-driven scenarios, enabling more flexible, zero-shot, and interactive use. Despite this progress, fine-grained video understanding remains challenging, particularly in cases involving subtle motions, ambiguous temporal cues, and small semantic differences between categories. Looking ahead, a key direction is to build universal video segmentation systems that scale to long videos and complex open-world concepts while remaining reliable and efficient in real-world applications.

\subsection{Video object tracking}
\label{sec:video_object_tracking}

Visual object tracking aims to localize a target object across video frames while coping with significant appearance variations caused by factors such as occlusion, deformation, motion blur, and illumination changes \cite{got10k,lasot,trackingnet}. Given an initial target specification, the goal is to continuously estimate the target state in subsequent frames with high accuracy and temporal consistency.
As illustrated in Fig.~\ref{fig:tracking}, the development of tracking methods can be viewed as a progression from pairwise appearance matching to temporal sequence modeling and further to multimodal representation learning.

\paragraph{RGB-based Tracking.}
Conventionally, most trackers follow a template–search paradigm \cite{sbt,sparsett,vital,atom,dimp}, where a target template extracted from the first frame (or dynamically updated over time) is matched against a search region \cite{siamcar,siamrpn,siamrpn++} in each new frame using correlation or similarity learning mechanisms.
With the rise of transformer architectures, recent works \cite{ostrack,seqtrack,kang2023exploring} increasingly adopt single-stream or joint modeling pipelines, where the template and search regions are processed together through self-attention mechanisms to enable global context modeling and long-range dependency capture. This paradigm has significantly advanced tracking performance by reducing hand-crafted design choices and allowing end-to-end learning of target representations and matching functions.

\newcommand{\NR}{\textcolor{gray}{NR}} 

\begin{table*}[t]
\centering
\setlength{\tabcolsep}{4pt}
\renewcommand{\arraystretch}{1.12}
\scriptsize

\centering
\small
\setlength{\tabcolsep}{5pt}
\renewcommand{\arraystretch}{1.1}
\resizebox{\textwidth}{!}{
\begin{tabular}{l l l l c c}
\toprule
\textbf{Method} &
\textbf{Backbone} &
\textbf{Initialization / Pre-training} &
\textbf{Input size} &
\textbf{Speed (FPS)} &
\textbf{Hardware} \\
\midrule

ViPT~\cite{zhu2023visual} &
\makecell[l]{ViT-B/16~\cite{dosovitskiy2020image}} &
\makecell[l]{OSTrack initialization} &
\makecell[l]{Template 128\\Search 256} &
24.78 &
A6000 \\

SDSTrack~\cite{hou2024sdstrack} &
\makecell[l]{ViT-B/16~\cite{dosovitskiy2020image}} &
\makecell[l]{OSTrack initialization} &
\makecell[l]{Template 128\\Search 256} &
20.86 &
A6000 \\

STTrack~\cite{hu2025exploiting} &
\makecell[l]{ViT-B/16~\cite{dosovitskiy2020image}} &
\makecell[l]{SOT foundation model} &
\makecell[l]{Template 128\\Search 256} &
— &
— \\

SUTrack~\cite{chen2024sutrack} &
\makecell[l]{HiViT-B~\cite{zhang2022hivit}} &
\makecell[l]{Fast-iTPN initialization (HiViT)\\+ CLIP-L text encoder} &
\makecell[l]{Template 112/192\\Search 224/384} &
12.0 &
RTX 2080Ti \\

FlexTrack~\cite{tan2025you} &
\makecell[l]{Video encoder\\+ HMoE fusion} &
\makecell[l]{Large-scale pretrained encoder} &
— &
— &
— \\

\bottomrule
\end{tabular}
}
\vspace{2pt}
\caption{\textbf{Key configurations and inference efficiency of representative multimodal trackers.} Reported FPS values are taken from the original papers and should be interpreted with care, as backbones, input resolutions, pre-training settings, and test hardware may differ across methods.}
\label{tab:tracking_eff}
\end{table*}

\paragraph{Multimodal Tracking.}
Despite these advances, the fundamental challenge of robustness to appearance changes remains largely unsolved \cite{zhao2025efficient,zhang2024multi}. Simply scaling model capacity or enlarging RGB training datasets is often insufficient, as many appearance variations stem from physical scene factors that are inherently ambiguous or invisible in RGB imagery alone. To address this limitation, some recent studies explore combining RGB trackers with language \cite{wang2021towards,feng2020real,hong2024onetracker}, leveraging their rich semantic priors and reasoning capabilities to assist high-level understanding and decision-making.

In parallel, another line of research directly tackles the visual ambiguity problem by incorporating additional sensing modalities \cite{yan2021depthtrack,wang2023visevent,li2021lasher}. Since appearance changes in tracking are often correlated with occlusion, fast motion, and adverse lighting conditions, there is growing interest in leveraging complementary sensors such as depth \cite{yang2022rgbd,he2021fast}, thermal \cite{zhang2022visible,zhao2021unified}, and event cameras \cite{zhang2022spiking,zhu2023cross}. These modalities provide alternative physical cues that are more invariant to specific failure modes of RGB, thereby offering stronger robustness in challenging environments.

\paragraph{Towards Unified and Robust Tracker.}

Early multimodal tracking approaches typically focused on single-modality-specific trackers \cite{zhang2022spiking,zhang2022visible,yang2022rgbd}, where separate models were designed and trained for each sensor configuration. More recently, the community has shifted toward unified tracking frameworks, which employ a single architecture \cite{zhu2023visual,protrack,hou2024sdstrack} and shared parameters \cite{untrack,tan2025you,tan2025xtrack} to handle multiple modalities. Such unified designs have been shown to encourage emergent cross-modal alignment, enabling the model to discover shared representations across different sensor inputs without explicit supervision.

Building upon this trend, recent works further extend unified trackers to support both RGB-only and RGB-X settings within a single framework \cite{chen2024sutrack}, allowing the same model to operate seamlessly \emph{with or without auxiliary modalities}. 
This naturally leads to increased attention on robustness under varying modality availability. 
Accordingly, we analyze the robustness of representative tracking models in Table~\ref{tab:tracking} by evaluating their performance under both the \textbf{Full} and \textbf{Miss} settings. The corresponding \texttt{*\textsubscript{miss}} benchmarks simulate real-world multi-sensor failures and temporary modality dropouts by introducing diverse missing patterns, including random missing, switched missing, and prolonged missing~\cite{tan2025you}.
Table~\ref{tab:tracking_eff} presents the key settings and efficiency of these methods. Among them, ViPT achieves the highest throughput at 24.78 FPS, followed by SDSTrack at 20.86 FPS and SUTrack at 12 FPS. ViPT and SDSTrack are the most directly comparable, as they share the same backbone, initialization, input resolution, and testing platform; under these matched settings, ViPT is more efficient. By contrast, SUTrack adopts a heavier multimodal design with additional encoders and larger input resolutions, which likely increases computational cost and reduces throughput. Overall, these results suggest that although multimodal tracking can improve representational capacity, its practical efficiency is still limited by the overhead of modality fusion and complex architectures.

\begin{table*}[!t]
    \centering
    \renewcommand{\arraystretch}{1.1}
    \resizebox{\textwidth}{!}{%
    \begin{tabular}{l l l l l}
        \toprule
        \textbf{Method} & \textbf{Paradigm} & \textbf{Architecture} & \textbf{Core Mechanism} & \textbf{Objective / Signal} \\
        \midrule
        \textbf{TGN}~\cite{chen2018temporally} & Supervised & 3D-CNN~\cite{tran2015learning} + GloVe~\cite{pennington2014glove} & Sequential Aggregation (LSTM/GRU) & Cross-Entropy Classification \\
        \textbf{ABLR}~\cite{yuan2019to} & Supervised & 3D-CNN + GloVe & Multimodal Cross-Attention & Smooth L1 Regression  \\
        \midrule
        \textbf{TGA}~\cite{mithun2019weakly} & Weakly Sup. & CLIP~\cite{radford2021learning} + GloVe & Text-Guided Attention & Reconstruction Loss \\
        \textbf{WSDEC}~\cite{duan2018weakly} & Weakly Sup. & Sentences localizer + generator & Dual-Task (Captioning + Localization) & Cycle-Consistency \\
        \midrule
        \textbf{TFVTG} ~\cite{zheng2024training} & Zero-Shot & BLIP-2~\cite{li2023blip} / CLIP & Sub-Event Decomposition & Logic-based Filtering \\
        \textbf{Moment-GPT}~\cite{xu2025zero} & Zero-Shot & Video-ChatGPT~\cite{Maaz2023VideoChatGPT} & Generative Scoring & Generation Confidence  \\
        \textbf{Time-R1}~\cite{wang2025time} & Post-Train (RL) & Qwen2.5-VL~\cite{bai2025qwen2} & Chain-of-Thought (CoT) & RL (IoU + Format Reward) \\
        \bottomrule
    \end{tabular}%
    }
    \vspace{5pt}
    \caption{\textbf{Comparative Analysis of Representative Video Temporal Grounding Methods.} The shift from TGN~\cite{chen2018temporally} to ABLR~\cite{yuan2019to} represents a shift from classification logic to regression logic while TGA~\cite{mithun2019weakly} and WSDEC~\cite{duan2018weakly} represent the supervision shift towards the weakly supervision manner. 
    Furthermore, later approaches increasingly intervene in the model’s internal reasoning process: TFVTG~\cite{zheng2024training} incorporates logic scripts, Moment-GPT~\cite{xu2025zero} leverages ensemble reasoning, and Time-R1~\cite{wang2025time} adopts reward-based optimization. 
    Overall, the field is moving toward deeper and more intrinsic control over model reasoning, rather than relying primarily on input manipulation or output-level heuristics. Results are taken from original papers, and experimental settings may differ; strict comparisons require reference to the original protocols.}
    \label{tab:vtg}
\end{table*}

To better handle long videos under online constraints, recent trackers~\cite{xiao2024mambatrack,ma2026smtrack} adopt selective state space models (SSMs) as a more efficient alternative to attention for long-range temporal modeling. 
For instance, SMTrack~\cite{ma2026smtrack} uses a state-aware Mamba design to propagate temporal information through hidden states. Such SSM-based trackers offer a lightweight and scalable way to preserve long-horizon temporal context, making them promising components for unified and robust tracking systems.

\paragraph{Summary}
Recent advances in video object tracking reflect a shift from purely appearance-based RGB matching toward multimodal and unified modeling.
An ideal RGB-X tracker should gracefully degrade to an RGB tracker when auxiliary sensors are absent, and conversely, function as an X-only tracker when RGB input is unreliable or missing. 
Existing studies demonstrate that achieving such robustness remains highly challenging, yet essential for real-world deployment, where sensor failure, occlusion, or bandwidth constraints are common. 
Consequently, designing a tracker that is accurate, modality-agnostic, and resilient to missing or degraded inputs continues to be a critical open problem in visual object tracking.

\subsection{Video temporal grounding}
\label{sec:video_grounding}

\begin{figure}[t]
    \setlength{\abovecaptionskip}{3mm}
    \centering
    \includegraphics[width=0.999\linewidth]{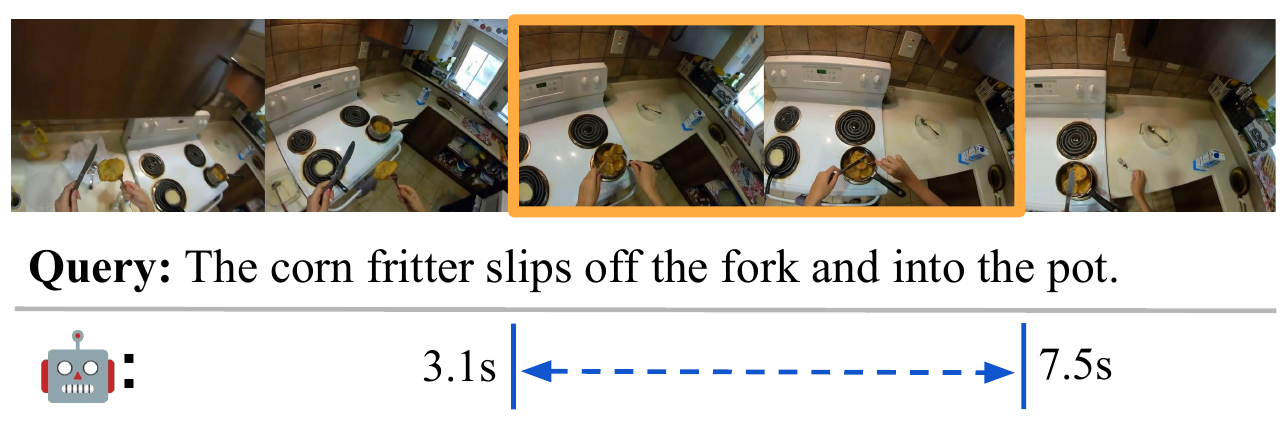}
\caption{\textbf{An illustration of video temporal grounding for the action.} The 10-second clip captures the transition of the fritter from being held by a fork to being successfully released into the cooking vessel. The highlighted frames demonstrate the model's ability to precisely localize the start and end points of this specific interaction.}
\label{fig:temporal_grounding}
\end{figure}

Video temporal grounding (VTG) aims to localize specific video segments that semantically correspond to a given natural language query~\cite{gao2017tall,anne2017localizing}. Unlike general video classification, VTG requires fine-grained temporal understanding to align complex linguistic concepts with dynamic visual content, as shown in Fig.~\ref{fig:temporal_grounding}. In this subsection, we organize prior work into three categories: \emph{before MLLMs}, \emph{supervised training MLLM-based methods}, and \emph{MLLM-based training-free methods}.
Table~\ref{tab:vtg} provides a comparative overview of representative VTG methods.

\paragraph{Before MLLMs.} Early VTG methods heavily relied on supervised pipelines, which were predominantly \emph{proposal-based}, treating grounding as a temporal object detection task~\cite{Xu2018TexttoClipVR,yuan2019semantic,chen2018temporally,zhang2019man,zhang2020learning}. Seminal works like TALL~\cite{gao2017tall} and MAN~\cite{zhang2019man} utilized sliding windows or anchors to generate candidates, followed by cross-modal matching. To mitigate the computational bottleneck of proposals, \emph{proposal-free} methods emerged, directly predicting probabilities or temporal boundaries at each time step~\cite{yuan2019to,ghosh2019excl,chen2019localizing,lu2019debug,zhang2020vslnet}. Parallel efforts in \emph{weakly supervised} learning sought to reduce annotation costs by leveraging video-level labels via reconstruction losses, multi-instance learning~\cite{mithun2019weakly,gao2019wslln,duan2018weakly,lin2020weakly}. Despite their effectiveness on specific benchmarks, these traditional models often struggle with open-vocabulary generalization and lack the intricate reasoning capabilities required for complex queries.

\paragraph{MLLM-based training methods.} Integrating MLLMs into VTG introduces powerful semantic reasoning and instruction-following capabilities. These methods typically follow two paradigms: \emph{pretraining} and \emph{fine-tuning}. Pretraining approaches aim to equip generalist MLLMs with native temporal awareness by training on large-scale time-annotated datasets. Representative frameworks like VTimeLLM~\cite{huang2024vtimellm} and TimeChat~\cite{ren2024timechat} employ multi-stage curricula, progressing from feature alignment to instruction tuning and boundary refinement, to transform the model into a temporal ``Executor'' capable of end-to-end localization. Time-R1~\cite{wang2025time} and VideoChat-R1~\cite{yan2025videochat} have also incorporated reinforcement learning to optimize non-differentiable metrics like IoU directly. Conversely, fine-tuning approaches adapt off-the-shelf MLLMs (\eg, BLIP-2~\cite{li2023blip}, LLaVA~\cite{liu2024visual}) to VTG tasks using smaller, domain-specific datasets. Models such as SeViLA~\cite{yu2023self} and LLaViLo~\cite{ma2023llavilo} leverage lightweight adapters or Q-Former designs to bridge the modality gap, learning to map video features to temporal tokens without the prohibitive cost of full pretraining. A particularly vital frontier within this domain is human-centric multimodal behavior cognition, where temporal instruction tuning and contrastive language-pose pretraining are increasingly employed to ground fine-grained human motions and repetitive actions in the wild ~\cite{li2025human,yao2025countllm,gu2025mocount}.

\paragraph{MLLM-based training-free methods.} Training-free methods leverage the zero-shot capabilities of pretrained MLLMs to perform grounding without parameter updates, prioritizing modularity and efficiency. These approaches generally fall into two strategies: \emph{feature similarity matching} and \emph{LLM-driven reasoning}. Similarity-based methods, such as Moment-GPT~\cite{xu2025zero} and TFVTG~\cite{zheng2024training}, extract semantic embeddings from queries and video segments using frozen encoders, localizing events by maximizing vector similarity. Reasoning-based methods, including Grounding-Prompter~\cite{chen2023grounding} and DeVi~\cite{qin2025question}, treat VTG as a textual inference task; they employ MLLMs to generate dense video captions or structured logs, which are then processed by an LLM (\eg, GPT-4~\cite{achiam2023gpt}) to reason about event boundaries logically. While these pipelines eliminate training costs, their precision is often bounded by the temporal resolution of the frozen features and the static context window of the LLM.

\paragraph{Summary.} Overall, video temporal grounding has evolved from hand-crafted proposal candidates or anchors to unified MLLM-based reasoning engines. While traditional methods established the foundations of boundary regression, MLLMs have revolutionized the field by enabling open-world understanding and complex causal reasoning. A central open challenge is to overcome the ``bag-of-features'' limitation of current vision encoders while supporting long-form temporal understanding and effectively integrating multimodal signals, including audio, visual, and text inputs, for precise grounding.
Another practical challenge is reducing model complexity. Recent studies~\cite{shen2024tempme,shen2025temporal} have begun to address this issue by removing redundant tokens and designing lightweight architectures for more efficient temporal grounding.

\begin{figure}[!t]
    \setlength{\abovecaptionskip}{3mm}
    \centering
    \includegraphics[width=0.999\linewidth]{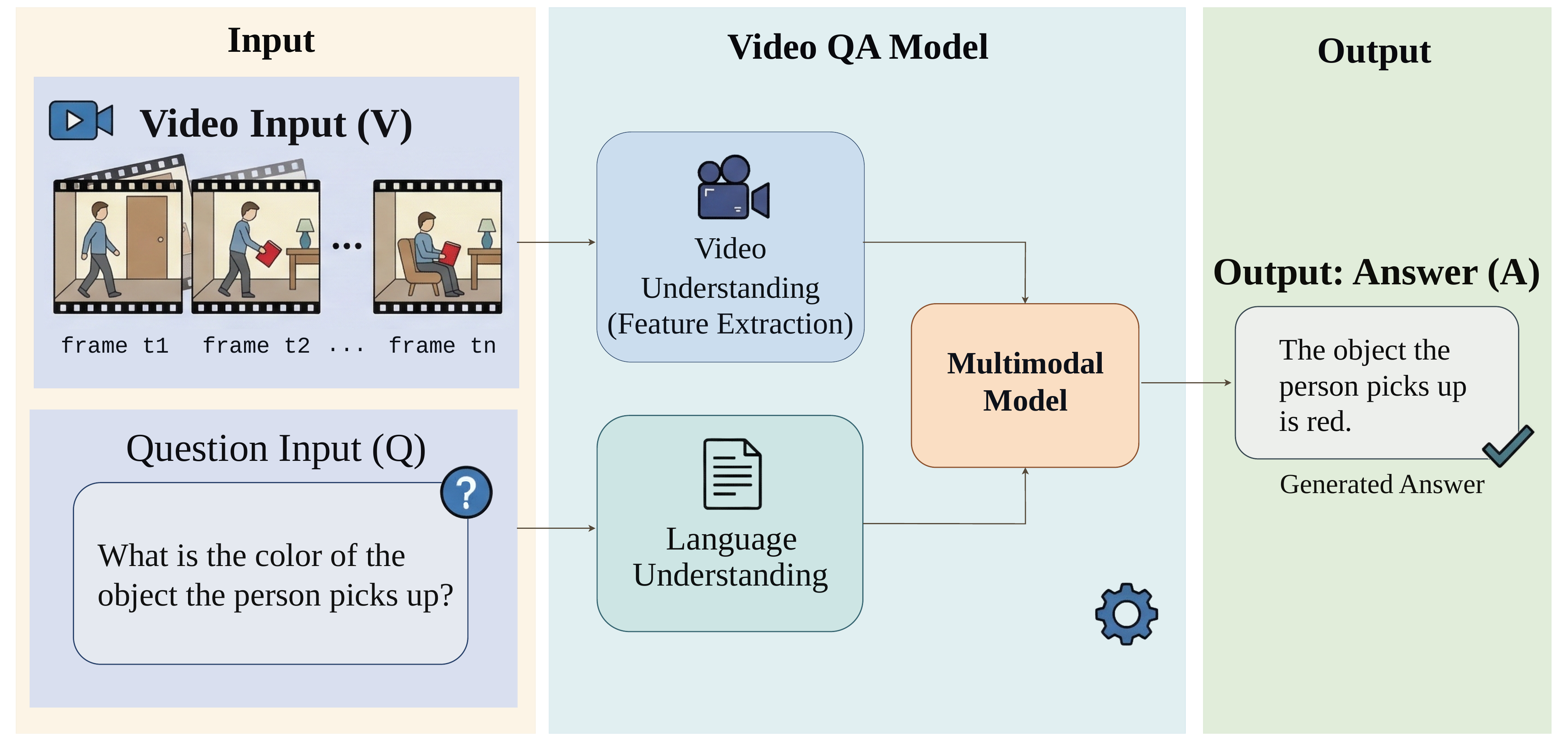}
\caption{\textbf{Video Question Answering.} The VideoQA task requires a model to jointly reason over visual and linguistic inputs. Given a video $V$ composed of temporal frames and a question $Q$, the system extracts video features and language representations, which are then fused within a multimodal model to generate a grounded natural language answer $A$. }
\label{fig:videoqa}
\end{figure}

\begin{table*}[t]
\centering
\resizebox{\textwidth}{!}{%
\begin{tabular}{@{}l L{5cm} p{9cm}@{}}
\toprule
\textbf{Category} & \textbf{Representative Works} & \textbf{Key Focus \& Technical Contributions} \\
\midrule
\multicolumn{3}{c}{\textit{\textbf{Evolution of Benchmarks}}} \\
\midrule
\textbf{Descriptive} & MSVD-QA~\cite{xu2017video}, MSRVTT-QA~\cite{xu2017video} & \textbf{Factoid Retrieval:} Recognition of atomic actions and objects in short clips limited by static bias. \\
\cmidrule(l){2-3}
\textbf{Temporal \& Causal} & TGIF-QA~\cite{Jang_2017_CVPR}, NExT-QA~\cite{xiao2021next}, ActivityNet-QA~\cite{yu2019activitynet} & \textbf{Relational Logic:} Tasks for repetition counting, state transitions, and causal interaction graphs like why and how. \\
\cmidrule(l){2-3}
\textbf{Long-Form Narrative} & EgoSchema~\cite{mangalam2023egoschema}, CinePile~\cite{rawal2024cinepile}, Video-MME~\cite{fu2025video}, LongVideoBench~\cite{wu2024longvideobench}, MLVU~\cite{zhou2024mlvu} & \textbf{Needle-in-a-Haystack:} Multi-hop reasoning over hour-scale contexts penalizing single-frame shortcuts via subset sensitivity. \\
\cmidrule(l){2-3}
\textbf{Spatial \& Active} & VSI-SUPER~\cite{yang2025cambrian}, StreamingCoT~\cite{hu2025streamingcot} & \textbf{World Modeling:} Tests long-horizon object permanence known as Spatial Supersensing and dynamic Chain-of-Thought. \\
\midrule
\multicolumn{3}{c}{\textit{\textbf{Architectural Paradigms}}} \\
\midrule
\textbf{Representation \& Alignment} & Video-LLaVA~\cite{lin2024video}, LLaMA-VID~\cite{li2024llama}, Chat-UniVi~\cite{jin2024chat}, Video-ChatGPT~\cite{Maaz2023VideoChatGPT}, LLaVA-OneVision~\cite{li2024llava} & \textbf{Tokenization:} Visual-language projection for zero-shot transfer and dual-token representation with context and content tokens. \\
\cmidrule(l){2-3}
\textbf{Resolution \& Scaling} & Qwen2-VL~\cite{wang2024qwen2}, InternVL 2~\cite{chen2024expanding}, LongVA~\cite{zhang2024long}, LongVU~\cite{shen2024longvu} & \textbf{Scaling and Long-video:} Naive Dynamic Resolution for arbitrary aspect ratios and massive parameter scaling up to 108B parameters. \\
\cmidrule(l){2-3}
\textbf{Modular \& Neuro-Symbolic} & MoReVQA~\cite{min2024morevqa}, NS-VideoQA~\cite{liang2024neural} & \textbf{Structured Reasoning:} Decomposition of queries into grounding and reasoning stages using symbolic logic for constraint adherence. \\
\cmidrule(l){2-3}
\textbf{Predictive Sensing} & Cambrian-S~\cite{yang2025cambrian} & \textbf{Active Inference:} Latent Frame Prediction using surprise metric for memory management and selective frame retention. \\
\bottomrule
\end{tabular}%
}
\vspace{5pt}
\caption{\textbf{Summary of VideoQA Benchmarks and Architectural Paradigms.} The field has evolved from short-term descriptive tasks to long-form, predictive, and spatial reasoning challenges.}
\label{tab:videoqa_summary}
\end{table*}

\section{Unified Video Understanding}
\label{sec:unified}
Beyond models that address video geometry and semantics in isolation, recent research increasingly explores \emph{unified} video understanding models that integrate low-level geometric understanding and high-level semantic cognition within a single framework. 
Such tasks require models to jointly capture physically grounded structure and semantically meaningful content, giving rise to the paradigm of \emph{unified video understanding}.
In this section, we review two representative directions of such unification: \emph{video question answering} (Section~\ref{sec:vqa}), which requires reasoning ability over geometric cues and semantic content to answer diverse geometry- and semantics-related queries, and \emph{unified video understanding and generation} (Section~\ref{sec:vug}), which further extends unification to video synthesis, enforcing the requirements of both physical consistency at the geometric level and semantic fidelity at the conceptual level.

\subsection{Video question answering}
\label{sec:vqa}
This section examines the transition of Video Question Answering (VideoQA) from static pattern recognition to dynamic, long-form reasoning.
 As shown in Fig.~\ref{fig:videoqa}, VideoQA requires a model to jointly process video frames and natural language questions to produce grounded answers.
Unlike ImageQA, which primarily tests spatial attribute binding, VideoQA demands the integration of spatio-temporal representations to interpret events as they unfold over time. Early approaches treated video as a bag of static frames, leading to ``static bias" where models could answer questions without processing temporal dynamics. Contemporary research has pivoted toward ``Cognitive Video Understanding", where recent benchmarks and Large Multimodal Models (LMMs) focus on causal inference, long-form narrative comprehension, and physical world modeling. A notable development in this domain is the concept of ``Spatial Supersensing," formalized by Cambrian-S, which posits that true video understanding requires a hierarchy of capabilities ranging from semantic perception to predictive world modeling. 
Table~\ref{tab:videoqa_summary} summarizes the literature by benchmarks and architectural paradigms.

\paragraph{Benchmarks.}
The benchmarking landscape has shifted from descriptive labeling to diagnostic tests of spatial memory and predictive capacity. The trajectory of VideoQA benchmarks has evolved from simple descriptive recognition in short clips, represented by MSVD-QA~\cite{xu2017video} and MSRVTT-QA~\cite{xu2017video}, to rigorous evaluations of temporal dynamics and causal logic in TGIF-QA~\cite{Jang_2017_CVPR}, NExT-QA~\cite{xiao2021next}, and ActivityNet-QA~\cite{yu2019activitynet}. This progression has culminated in a focus on long-form narrative understanding, where EgoSchema~\cite{mangalam2023egoschema}, CinePile~\cite{rawal2024cinepile}, Video-MME~\cite{fu2025video}, LongVideoBench~\cite{wu2024longvideobench}, and MLVU~\cite{zhou2024mlvu} challenge models to retrieve ``needle-in-a-haystack" details and perform multi-hop reasoning across hour-scale contexts. Pushing beyond passive perception, the latest wave of benchmarks targets ``Spatial Supersensing" and active world modeling: VSI-SUPER~\cite{yang2025cambrian}  tests long-horizon object permanence, StreamingCoT~\cite{hu2025streamingcot} evaluates explicit dynamic Chain-of-Thought reasoning.

\paragraph{Architectural Paradigms.}
Current architectural trends have bifurcated into foundation models that scale context windows, native resolution specialists, and modular neuro-symbolic systems. In the realm of Multimodal Large Language Models, to effectively incorporate video modality, Video-LLaVA~\cite{lin2024video}  aligns visual features into the language space prior to projection to enable zero-shot generalization, LLaMA-VID~\cite{li2024llama}  mitigates the cost of long videos via a dual-token representation strategy, Chat-UniVi~\cite{jin2024chat} utilizes dynamic visual tokens to efficiently model spatiotemporal dependencies, Video-ChatGPT~\cite{Maaz2023VideoChatGPT} establishes a baseline for conversational video understanding with quantitative evaluation frameworks, and LLaVA-OneVision~\cite{li2024llava} improves temporal coherence through training on high-quality synthetic data. Addressing resolution and context bottlenecks, Qwen2-VL~\cite{wang2024qwen2} integrates a ``Naive Dynamic Resolution" mechanism for arbitrary aspect ratios, InternVL 2~\cite{chen2024expanding} validates scaling laws with a massive 108B parameter architecture, LongVA~\cite{zhang2024long} extends context capabilities to process up to 2000 frames, and LongVU~\cite{shen2024longvu} employs spatiotemporal adaptive compression to maintain high fidelity in long contexts. Prioritizing interpretability and causal logic, MoReVQA~\cite{min2024morevqa} decomposes queries into grounding and reasoning stages, NS-VideoQA~\cite{liang2024neural} combines neural perception with symbolic logic programming for strict constraint adherence, and Cambrian-S~\cite{yang2025cambrian} pioneers ``Predictive Sensing" using a self-supervised Latent Frame Prediction module to manage memory by selectively retaining frames.

Another line of work~\cite{xu2025timeviper,ren2025vamba} adopts hybrid SSM--Transformer designs, using state-space modeling for long-context processing while retaining attention for selective high-bandwidth interactions. 
For example, TimeViper~\cite{xu2025timeviper} combines Mamba and Transformer blocks with a token-transfer mechanism to reduce visual token redundancy, while Vamba~\cite{ren2025vamba} uses Mamba-2 blocks to encode long video sequences with linear complexity. Overall, these hybrid designs alleviate the quadratic cost of attention and improve memory scalability for long-video understanding.

\begin{figure*}[t]
    \centering
    \includegraphics[width=\textwidth]{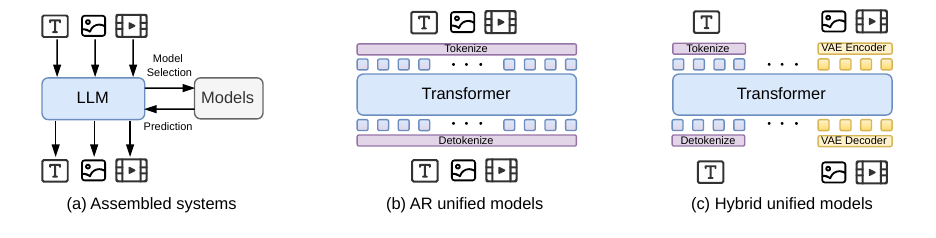} 
    \caption{\textbf{Different framework designs of unified video understanding and generation models}: (a) Assembled systems rely on an LLM to orchestrate external video expert models. (b) Autoregressive (AR) unified models map text, image, and video into a single token stream, processing them through a unified transformer with a discrete next-token prediction objective. (c) Hybrid unified models utilize a shared transformer backbone but incorporate continuous generation targets (\eg, diffusion or flow matching) via VAE decoders to better match the demands of video synthesis.}
    \label{fig:unified_design_space}
    \vspace{-8pt}
\end{figure*}

\paragraph{Discussion.}
In summary, VideoQA is undergoing a paradigm shift from passive visual retrieval to active, predictive cognition. While LMMs have demonstrated impressive generalization, recent benchmarks reveal that simply scaling context windows faces diminishing returns in maintaining spatial consistency and object permanence over unlimited streams. Indeed, investigating the intrinsic entropy of context length scaling and the specific mechanisms of deductive versus inductive reasoning within LLMs has become essential for understanding these fundamental limitations and achieving reliable attention consistency ~\cite{shi2026entropy,cai2025reasoning,lan2025attention}. Future research is increasingly converging on World Models that integrate predictive sensing to actively filter and organize visual memories rather than passively storing them. Furthermore, to address the hallucinations common in generative models, there is a critical need to integrate structured reasoning that enforces temporal logic and causal fidelity, paving the way for agents capable of continuous, embodied interaction.

\begin{table*}[t]
\centering
\small
\setlength{\tabcolsep}{9pt}
\renewcommand{\arraystretch}{1.1}
\resizebox{\textwidth}{!}{
\begin{tabular}{l c c l c c}
\hline
\textbf{Model} & \textbf{Venue} & \textbf{Year} & \textbf{Backbone} & \textbf{Unified} & \textbf{Paradigm} \\
\hline
HuggingGPT~\cite{shen2023hugginggpt} & NeurIPS & 2023 & GPT-3.5 & Assembled & Tool \\

NExT-GPT~\cite{wu2024next} & ICML & 2024 & Vicuna~\cite{chiang2023vicuna} & Assembled & AR* \\
Video-LaVIT~\cite{jin2024video} & ACL & 2024 & LLaMA-2~\cite{touvron2023llama} & Native & AR \\
Emu3~\cite{wang2024emu3} & arXiv & 2024 & LLaMA-2 & Native & AR \\

VILA-U~\cite{wu2024vila} & ICLR & 2025 & LLaMA-2 & Native & AR \\
BAGEL~\cite{deng2025emerging} & arXiv & 2025 & Qwen-2.5~\cite{yang2025qwen3} & Native & AR+Flow \\
HaploOmni~\cite{xiao2025haploomni} & arXiv & 2025 & Qwen-2.5+CogVideoX~\cite{yang2024cogvideox} & Native & AR+Diff \\
Show-o2~\cite{xie2025showo2} & NeurIPS & 2025 & Qwen-2.5 & Native & AR \\
Omni-Video~\cite{tan2025omni} & arXiv & 2025 & VILA~\cite{Lin_2024_CVPR} & Assembled & AR* \\
UniVid~\cite{luo2025univid} & arXiv & 2025 & BAGEL-7B~\cite{deng2025emerging}+Wan2.2~\cite{wan2025wan} & Assembled & AR+Diff \\
UniVideo~\cite{wei2025univideo} & ICLR & 2026 & Qwen-2.5+HunyuanVideo~\cite{kong2024hunyuanvideo} & Assembled & AR+Diff \\
TUNA~\cite{liu2025tuna} & CVPR & 2026 & Qwen-2.5 & Native & AR+Flow \\
\hline
\end{tabular}
}
\vspace{5pt}
\caption{\textbf{Unified video understanding and generation models.}
``Tool'' denotes LLM-orchestrated external expert models (not end-to-end).
``*'' indicates external generation modules (not end-to-end).
Paradigms: autoregressive (AR), diffusion (Diff), and flow matching (Flow).}

\label{tab:unified_video_umms}
\end{table*}

\subsection{Video understanding and generation}
\label{sec:vug}

Unified Multimodal Models (UMMs) aim to build a single system that supports both multimodal \emph{understanding} (\eg, QA, grounding, reasoning) and visual \emph{generation} (\eg, synthesis and editing), and have shown strong progress in the image domain and beyond (including unified and tool-augmented multimodal systems)~\cite{liu2025worldweaver,huang2026vecglypher,liu2023cones,liu2023customizable,sun2024generative,xie2024showo,HaploVL,tong2025metamorph,yang2025mmada,ji2025colon,li2025dreambeast}. 
More recently, a growing body of work has extended this paradigm to videos, spanning video understanding, video generation, and instruction-based video editing~\cite{wu2024vila,wang2024emu3,xie2025showo2,liu2025tuna,tan2025omni,luo2025univid,wei2025univideo,mou2025instructx,lu2025uniugp}. 
Table~\ref{tab:unified_video_umms} summarizes representative unified video models and highlights how different systems realize unification across architectures and training paradigms.
Compared with images, video unification must handle longer temporal context, stricter spatiotemporal consistency, and substantially higher compute, while still preserving language alignment and controllability. We group video-oriented UMMs into three representative families: \textbf{assembled systems}, \textbf{autoregressive (AR) unified models}, and \textbf{hybrid unified models}. Fig. \ref{fig:unified_design_space} illustrates the design space of these three paradigms, highlighting their structural differences in integrating multimodal understanding and generation.

\paragraph{Assembled systems.}
A practical route to ``unified'' capability is to place an LLM in charge of calling external expert models for captioning, generation, or editing, and then combining their outputs into a single response. HuggingGPT~\cite{shen2023hugginggpt} represents this tool-based philosophy, and video-oriented extensions follow the same recipe by pairing strong instruction following with specialized video modules~\cite{wu2024next,tan2025omni,luo2025univid,wei2025univideo}. This family is attractive because it is modular: stronger video generators or editors can be plugged in without retraining the entire system. The downside is that the system is not optimized end-to-end for video synthesis, and temporal consistency can depend heavily on the handoff between components.

\paragraph{Autoregressive unified models.}
A conceptually clean unification strategy is to map text, images, and videos into a single token stream and train a decoder-only Transformer with one AR next-token objective. Video-LaVIT~\cite{jin2024video} and VILA-U~\cite{wu2024vila} are representative AR-style video UMMs, while Emu3~\cite{wang2024emu3} further strengthens the ``one model, one objective'' view by generating discrete visual tokens. The appeal is objective-level unification: the same next-token prediction recipe supports both understanding (generate text conditioned on video) and generation (generate video conditioned on text). The main bottleneck is the length--fidelity trade-off. High-quality video either demands long sequences or expressive tokenizers, both of which stress context length, memory, and training stability. As a result, AR unified models must balance efficiency, temporal fidelity, and instruction-grounded controllability.

\paragraph{Hybrid unified models.}
A third family keeps a unified backbone and interface, but relies on diffusion- or flow-style objectives to better match the demands of video synthesis. Show-o2~\cite{xie2025showo2}, TUNA~\cite{liu2025tuna}, and BAGEL~\cite{deng2025emerging} represent hybrid ``native'' unification that blends language-model-like decoding with iterative refinement, aiming to improve visual quality while keeping a shared representation. In parallel, several video systems combine AR-style instruction following with diffusion-based generators and editing modules, delivering strong editing ability and high synthesis quality in practice~\cite{luo2025univid,xiao2025haploomni,wei2025univideo,mou2025instructx}.
Beyond modeling design, recent work has improved the efficiency of training and deploying unified models through diffusion-based weight generation, in-context meta-LoRA generation, and hybrid-policy optimization~\cite{guan2026mcdi,guan2025lohp,shao2025icmlora}. 
Compared with pure AR objective, hybrid designs can improve realism and temporal coherence, but add new training challenges, including how to share capacity between understanding and generation, and how to maintain instruction faithfulness while preserving consistency over long horizons.

\paragraph{Discussion.}
Across these three families, unification offers a single interface for understanding, generation, and editing. Assembled systems prioritize practicality and modular upgrades; AR unified models prioritize simplicity and a single training objective; hybrid models try to keep a unified representation while achieving higher-fidelity video synthesis.
Despite this promise, unified modeling still faces important limitations, including degradation over long horizons, sensitivity to prompts, and imbalanced performance across understanding and generation tasks.

\section{Conclusion and Outlook}
\label{sec:conclusion}
This survey reviewed recent progress in video understanding from the perspectives of low-level geometric perception, high-level semantic reasoning, and emerging unified modeling paradigms. 
We first examined advances in low-level geometry understanding, highlighting a convergence toward joint feed-forward geometry models that enable increasingly accurate and efficient recovery of 3D structure and motion from videos. 
We then surveyed high-level semantic understanding tasks such as segmentation, tracking, and temporal grounding, which place strong demands on spatiotemporal reasoning and cross-modal alignment for videos. 
Finally, we discussed unified video understanding models that bridge geometry, semantics, and generation within a single framework, reflecting a broader shift toward holistic video foundation models.

Looking forward, several challenges and opportunities stand out.

\textbf{First, toward world models with active and predictive understanding.}
A clear trend is the emergence of a \emph{world model} paradigm, in which understanding across tasks and abstraction levels is progressively integrated into a single model capable of both \emph{perceiving current observations and predicting future states}.
The ambition to unify diverse visual tasks echoes earlier efforts in vision~\cite{zamir2018taskonomy,mizrahi20234m,bachmann20244m}, which explored shared representations across multiple tasks in \emph{static images}; extending this philosophy to video introduces additional challenges and opportunities.
In parallel, rather than passively processing visual inputs, emerging video models are expected to actively interpret and anticipate scene evolution, with generation serving as a mechanism for \emph{zero-shot} problem solving and counterfactual reasoning~\cite{wiedemer2025video}.
Achieving this requires tighter integration between geometry and semantics. While recent models demonstrate promising joint representations, fully leveraging geometric constraints to stabilize semantic reasoning, and conversely using semantic cues to guide geometry in dynamic and ambiguous scenes, remains an open problem.
Moreover, unification with generation introduces new demands on physical plausibility and instruction faithfulness, calling for stronger predictive sensing and grounding mechanisms to mitigate hallucination and drift.

\textbf{Second, memory as a first-class design principle for video models.}
At its core, video understanding is fundamentally \emph{a problem of memory}.
As models move toward processing hour-scale videos or effectively unbounded real-world streams, a central challenge is how to design memory mechanisms that balance latency, computational cost, and representational fidelity. 
Future systems must maintain and update both an internal model state and an external scene state, \emph{selectively retaining relevant information while discarding redundancy}. 
This calls for advances in memory-efficient architectures, streaming and causal inference, and adaptive state representations that go beyond naïve context expansion~\cite{an2025onestory,yu2025context,qiu2025histream}, enabling scalable and persistent world modeling.
At the same time, it remains unclear how much of video understanding truly requires long-horizon memory, versus what can already be achieved with limited temporal cues, such as short frame bursts or motion signals like optical flow.
In analogy to David Marr’s sketches from 2D to 2.5D representations~\cite{marr1978representation}, such intermediate regimes can unlock substantially richer geometric and semantic reasoning than single-frame inputs, without requiring full long-horizon temporal modeling.
Viewed through this lens, they highlight memory granularity as a design dimension and offer a principled trade-off between representational richness and memory efficiency for addressing video understanding problems along this spectrum.

\textbf{Third, uncertainty-aware planning and decision-making from video understanding.}
Beyond deterministic objectives such as video labeling, an open question is how far video understanding should extend toward understanding \emph{future uncertainty}. 
While some scenarios, such as physical simulation or reconstruction, demand highly accurate and deterministic prediction, the real world evolves in fundamentally uncertain ways. 
In such settings, understanding is valuable not only for prediction accuracy, but also for \emph{decision-making}: models must reason over multiple plausible futures, evaluate trade-offs between goals and costs, and support planning under uncertainty. 
This perspective naturally connects video understanding with \emph{embodied and agent-centric settings}, where models operate in an online or streaming manner and must continuously interpret visual observations to guide actions. 
Developing such models would move video understanding beyond recognition and prediction toward deeper capabilities for reasoning, exploration, and interaction in dynamic environments.

More broadly, the field is transitioning from isolated task performance toward \emph{systems that reason, predict, and act over time}.
Unified video understanding models provide a practical lens on this transition, as they naturally expose both the synergies and tensions between perception, reasoning, and synthesis.
We expect continued progress along these directions to play a critical role in enabling reliable, general-purpose video agents for real-world applications.

\vspace{0.5em}

\bibliographystyle{IEEEtran} 
\bibliography{camera_ready/reference-camera-ready}       

\section*{Acknowledgments}
Zhaochong An, Jiaang Li, and Serge Belongie are supported by funding from the Pioneer Centre for AI, DNRF grant number P1.



\noindent
\newpage

\vspace{20pt}
\noindent{\bf Zhaochong An} is a PhD student at the University of Copenhagen, affiliated with Pioneer Centre for Artificial Intelligence, under the ELLlS program. He is advised by Prof. Serge Belongie.
He received his MSc in Computer Science from ETH Zurich in 2023 under the supervision of Prof. Luc Van Gool. His research primarily focuses on computer vision and deep learning, including 3D understanding, video generation, and multimodal models.

E-mail: zhan@di.ku.dk

ORCID iD: 0009-0007-5985-7470

\noindent{\bf Zirui Li}
is expected to receive the M.S. degree from Beijing University of Post and Telecommunication, Beijing, China, in 2026, and will pursue the Ph.D. degree at Nankai University, Tianjin, China, advised by Prof. Guolei Sun. His current research interests include object tracking and multimodal large language models for video understanding.

E-mail: lizirui2019@bupt.edu.cn

\noindent{\bf Mingqiao Ye}
is a computer science PhD student at EPFL. He is advised by Prof. Amir Zamir. He received his MSc from ETH Zurich in 2024. His research primarily focuses on multimodal foundation models, including pre-training and post-training.

Email: mingqiao.ye@epfl.chs

\noindent{\bf Feng Qiao} is a computer science PhD student at Washington University in St. Louis. He is advised by Prof. Nathan Jacobs. He received his MSc from RWTH-Aachen in 2019. His research primarily focuses on computer vision and deep learning, including 3D Vision, video generation, and autonomous driving.

E-mail: f.qiao@wustl.edu

\noindent{\bf Jiaang Li}
is a PhD student advised by Prof. Serge Belongie at the University of Copenhagen, affiliated with Pioneer Centre for Artificial Intelligence, under the ELLlS program. He received his MSc in Computer Science from University of Copenhagen in 2023 under the supervision of Prof. Anders S{\o}gaard. His research primarily focuses on multimodal alignment and transferring with MLLMs.

Email: jili@di.ku.dk

\noindent{\bf Zongwei Wu}
is a Postdoctoral Associate Researcher at the University of Wuerzburg and member of ELLIS. He held a Postdoctoral position at the University of Burgundy (CNRS-ICB), France and obtained Ph.D. from Université Bourgogne Franche-Comté (CNRS-Vibot), France. His research primarily focuses on multisensor fusion.

Email: zongwei.wu@uni-wuerzburg.de

\noindent{\bf Vishal Thengane} is a joint PhD student at the University of Surrey (UoS), United Kingdom, and the University of Wollongong (UoW), Australia, focusing on resource- and data-efficient methods for scene understanding. His research is supervised by Xiatian Zhu and Lu Yin at UoS, and by Salim Bouzerdoum and Son Lam Phung at UoW.

Email: vgthengane@gmail.com

\noindent{\bf Chengzu Li}
is a PhD student at the Language Technology Lab, Unversity of Cambridge, affiliated with Pioneer Center for Artificial Intelligence. He is advised by Prof. Anna Korhonen, Dr. Ivan Vulić and Prof. Serge Belongie. He received his MPhil in Advanced Computer Science from University of Cambridge in 2023 supervised by Prof. Simone Teufel. His research primarily focuses on multimodal reasoning with MLLMs and generative models. 

Email: cl917@cam.ac.uk

\noindent{\bf Lei Li}
is currently a full professor at the School of Artificial Intelligence, Beijing Institute of Technology. He received his Ph.D. degree from the University of Copenhagen. He was a researcher at the University of Washington and University of Copenhagen, associated with the Pioneer Centre for AI. He focuses on Multimodal Learning, Generative AI, and Optimization.

Email: lenny.lilei.cs@gmail.com

\noindent{\bf Luc Van Gool} is currently a full professor of computer vision with INSAIT, Sofia University St. Kliment Ohridski, and professor emeritus with ETH Zürich and the KU Leuven. His research interests include 2D and 3D object recognition, texture analysis, range acquisition, stereo vision, robot vision, and optical flow. He has led research on autonomous cars in the context of the Toyota TRACE Labs with ETH and in Leuven, and has an extensive collaboration with Huawei on the topic of image and video enhancement. He was the recipient of the several Best Paper awards (e.g. David Marr Prize 1998, Best Paper CVPR 2007), the Koenderink Award in 2016, and the “Distinguished Researcher” nomination by the IEEE Computer Society in 2017. He was the holder of an ERC Advanced Grant. He is also a program committee member of several major computer vision conferences.

E-mail: vangool@vision.ee.ethz.ch

\noindent{\bf Guolei Sun} 
is currently a full professor at Nankai University. He received master degree from King Abdullah University of Science and Technology in 2018 and doctoral degree from ETH Zurich in 2024. From 2024 to 2025, he worked as a postdoctoral researcher in ETH Zurich. His research interests lie in computer vision and deep learning for tasks such as video understanding, and multi-modal large language models. He has published more than 40 papers in top journals and conferences such as TPAMI, CVPR, ICCV, and ECCV. He also serves as an area chair for ICLR.

E-mail: guolei.sun@nankai.edu.cn

ORCID iD: 0000-0001-8667-9656

\noindent{\bf Serge Belongie} is a professor of Computer Science at the University of Copenhagen, where he also serves as the head of the Pioneer Centre for Artificial Intelligence (P1). Previously, he was a professor of Computer Science at Cornell University, an Associate Dean at Cornell Tech, a member of the Visiting Faculty program at Google, and a professor of Computer Science \& Engineering at UC San Diego. His research interests include Computer Vision, Machine Learning, and Human-in-the-Loop Computing. He is also a co-founder of several companies including Digital Persona and Anchovi Labs. He is a recipient of the NSF CAREER Award, the Alfred P.~Sloan Research Fellowship, the MIT Technology Review ``Innovators Under 35'' Award, the Stibitz-Wilson Award, the Helmholtz Prize, the Everingham Prize, and the Koenderink Prize for fundamental contributions to the Computer Vision community. He is a member of the Royal Danish Academy of Sciences and Letters and serves as president of the board of the European Laboratory for Learning and Intelligent Systems (ELLIS).

E-mail: s.belongie@di.ku.dk

\end{document}